\pdfoutput=1

\documentclass[11pt]{article}

\usepackage[final]{acl}

\usepackage{times}
\usepackage{latexsym}

\usepackage[T1]{fontenc}

\usepackage[utf8]{inputenc}

\usepackage{microtype}

\usepackage{inconsolata}

\usepackage{graphicx}

\usepackage{fancybox}
\usepackage{multirow}   
\usepackage{booktabs}
\usepackage{tabularx}
\usepackage{xcolor}
\usepackage{amsmath}
\usepackage{colortbl}
\usepackage{subcaption}

\newcommand{\cd}{CCD}

\newcommand{\context}{{\scshape Context}}
\newcommand{\abstain}{{\scshape Abstain}}
\newcommand{\sa}{{\scshape Self-Ask}}
\newcommand{\cad}{{\scshape CAD}}
\newcommand{\acd}{{\scshape ACD}}
\newcommand{\fsb}{{\scshape FSB}}
\newcommand{\acda}{{\scshape ACD-A}}
\newcommand{\ent}{{\scshape Entropy}}
\newcommand{\method}{{\scshape CDA}}
\newcommand{\methodm}{{\scshape CDA-m}}
\newcommand{\both}{{\scshape CDA(-m)}}

\newcommand{\1}{N$_{\text{1}}$}
\newcommand{\2}{N$_{\text{2}}$}
\newcommand{\3}{N$_{\text{3}}$}
\newcommand{\4}{N$_{\text{4}}$}
\newcommand{\5}{N$_{\text{5}}$}
\newcommand{\full}{N}

\newcommand{\ml}{{\scshape Llama3 8B Instruct}}
\newcommand{\mll}{{\scshape Llama2 7B Chat}}
\newcommand{\mlll}{{\scshape Llama2 13B Chat}}
\newcommand{\mm}{{\scshape Mistral 7B Instruct}}

\newcommand{\abs}{F1$_{\text{abs}}$}
\newcommand{\ans}{F1$_{\text{ans}}$}

\newcommand{\nullprompt}{\texttt{null}}

\newcommand{\entropy}{$\mathcal{H}$}

\newcommand{\pp}{$\mathcal{P}$}
\newcommand{\cc}{$\mathcal{C}$}

\newcommand{\astfootnote}[1]{
    \let\oldthefootnote=\thefootnote
    \setcounter{footnote}{1}
    \renewcommand{\thefootnote}{\fnsymbol{footnote}}
    \footnotetext{#1}
    \let\thefootnote=\oldthefootnote
}

\title{When to Speak, When to Abstain: Contrastive Decoding with Abstention}

\author{
    Hyuhng Joon Kim\textsuperscript{\rm 1},
    Youna Kim\textsuperscript{\rm 1},
    Sang-goo Lee\textsuperscript{\rm 1 2},
    Taeuk Kim\textsuperscript{\rm 3 *}\\
    \textsuperscript{\rm 1}Seoul National University,
    \textsuperscript{\rm 2}IntelliSys, Korea,
    \textsuperscript{\rm 3}Hanyang University\\
    \{heyjoonkim, anna9812, sglee\}@europa.snu.ac.kr\\
    kimtaeuk@hanyang.ac.kr
}

\begin{document}
\maketitle



\begin{abstract}
Large Language Models (LLMs) demonstrate exceptional performance across diverse tasks by leveraging pre-trained (\textit{i.e., parametric}) and external (\textit{i.e., contextual}) knowledge.
While substantial efforts have been made to enhance the utilization of both forms of knowledge, situations in which models lack relevant information remain underexplored.
To investigate this challenge, we first present a controlled testbed featuring four distinct knowledge access scenarios, including the aforementioned edge case, revealing that conventional LLM usage exhibits insufficient robustness in handling all instances.
Addressing this limitation, we propose \textbf{Contrastive Decoding with Abstention (\method{})}, a novel training-free decoding method that allows LLMs to generate responses when relevant knowledge is available and to abstain otherwise.
\method{} estimates the relevance of both knowledge sources for a given input, adaptively deciding which type of information to prioritize and which to exclude.
Through extensive experiments, we demonstrate that \method{} can effectively perform accurate generation and abstention simultaneously, enhancing reliability and preserving user trust.
\end{abstract}
\astfootnote{Corresponding author.}



\section{Introduction}

\begin{figure}[t!]
    \centering
    
    \includegraphics[width=0.97\columnwidth]{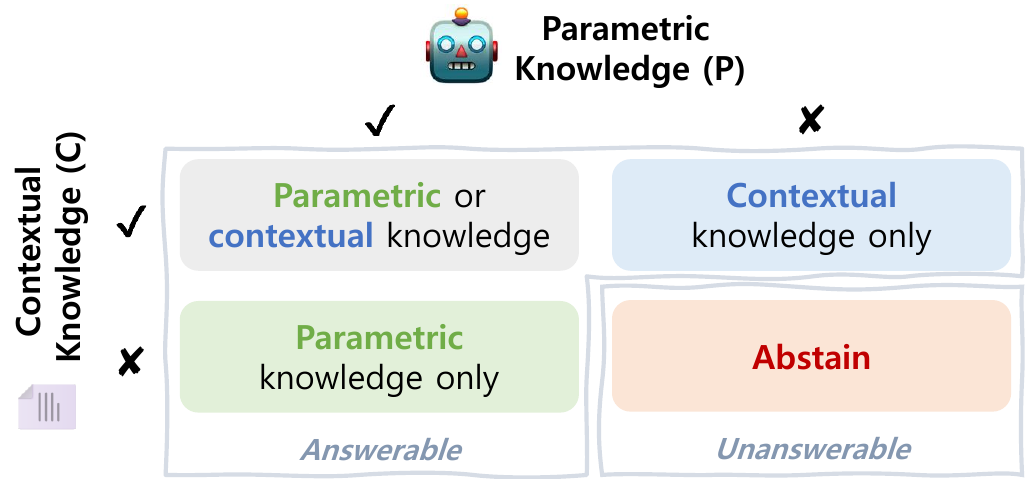}

    \caption{
        This study considers four possible scenarios based on the existence of the model's parametric and contextual knowledge. 
        The model is expected to respond \textit{reliably} by either (1) generating correct responses leveraging any form of relevant knowledge available or (2) abstaining from producing potentially inaccurate or misleading outputs when no relevant knowledge exists.
    }
    \label{figure:intro_figure}
\end{figure}

Large Language Models (LLMs) \citep{team2023gemini,achiam2023gpt,dubey2024llama} acquire extensive \textit{parametric knowledge} during pre-training, which helps them attain remarkable performance across a range of tasks.
However, no matter how comprehensive and useful parametric knowledge may be, it is inherently limited by the scope of the pre-training corpus.
Consequently, LLMs become less reliable when processing inputs from underrepresented domains, such as those including domain-specific \citep{10.5555/3618408.3619049,raja2024rag,feng-etal-2024-retrieval} or outdated data \citep{10.5555/3540261.3542508,kasai2023realtime,zhao-etal-2024-set}.

To overcome this challenge, approaches that integrate previously unseen information during inference have emerged \citep{buttcher2016information,10.1145/2939672.2939677,karpukhin-etal-2020-dense}.
They provide external information to LLMs as \textit{contextual knowledge}, 
expanding the knowledge boundary beyond what is learned from training.

Since LLMs are generally exposed to two distinct sources of information---\textit{parametric} and \textit{contextual} knowledge---they are expected to adaptively leverage both to maximize performance.
Despite efforts to enhance such desired behavior, scenarios where neither parametric nor contextual knowledge is available—often encountered in real-world settings—remain largely underexplored.
Compelling models to respond imprudently in such cases heightens the risk of hallucination, diminishes reliability, and introduces potential dangers in high-stakes applications.

Therefore, it is crucial for LLMs to abstain from responding when necessary information is inaccessible \citep{varshney-etal-2024-art,zhang-etal-2024-r,wen2024knowlimitssurveyabstention} while preserving performance when relevant knowledge is available.
However, such behavior requires a precise assessment of the knowledge and the integration of this assessment into the generation process, both of which are inherently challenging.

In this work, we first present a controlled testbed, where the accessibility of both types of knowledge for a query is explicitly determined.
In contrast to typical scenarios where definitively confirming the availability of parametric or contextual knowledge is imprecise, our experimental setup facilitates controlled investigations, covering all scenarios depicted in Figure \ref{figure:intro_figure}.
Experimental results on this testbed indicate that existing methods for LLMs lack sufficient robustness in effectively handling all the considered scenarios.

To this end, we propose \textbf{Contrastive Decoding with Abstention (\method{})}, a novel, training-free decoding method that enables LLMs to not only leverage relevant parametric or contextual knowledge during generation but also abstain when no appropriate knowledge is available.
During the decoding process, \method{} assesses the relevance of both forms of knowledge, 
adaptively determining the knowledge to attend during generation.
Moreover, \method{} steers the models towards abstention if no relevant knowledge is available.
The relevancy is estimated as the uncertainty associated with the knowledge in response to a specific query.


Extensive experiments with four LLMs on three question-answering (QA) datasets \citep{ZHANG20231,etezadi2023state} demonstrate that \method{} effectively enables LLMs to abstain in the absence of relevant knowledge while maintaining existing capabilities without additional training.
Further validation against training-based methods demonstrated \method{}'s robust generalization capabilities, while evaluations in retrieval-augmented generation (RAG) setting highlight its effectiveness across practical scenarios.




\section{Related Work}
\subsection{Contrastive Decoding}
Contrastive decoding (CD) controls text generation by contrasting different output distributions and steers the model in the desired direction. 
DExperts \citep{liu-etal-2021-dexperts} employs an ensemble of an “expert” and an “anti-expert” for tasks such as detoxification.
\citet{li-etal-2023-contrastive} contrasts the output distributions of a large LM and a small LM for open-ended text generation.
CD is also proven effective in domains such as reasoning \citep{o2023contrastive} and machine translation \citep{waldendorf-etal-2024-contrastive}. 
Recently, there has been growing interest in context-aware contrastive decoding (\cd{}) \citep{zhao-etal-2024-enhancing,kim-etal-2024-adaptive,qiu2024entropy,shi-etal-2024-trusting}, which enables the model to leverage both parametric and contextual knowledge during decoding, tackling tasks such as knowledge conflicts \citep{longpre-etal-2021-entity,chen-etal-2022-rich,zhou-etal-2023-context}.
Despite the promising results, existing approaches assume at least one knowledge source is always available.
In practice, LLMs frequently encounter situations with no relevant knowledge, a gap these methods fail to bridge.
To address this limitation, we expand the scope to include such edge cases and propose a novel approach of integrating abstention to \cd{}. 

\subsection{Abstention in LLMs}
LLMs often generate unintended or undesirable responses, such as hallucinations \citep{maynez-etal-2020-faithfulness,10.1145/3571730,jiang-etal-2024-large}, 
biases \citep{sap-etal-2020-social,feng-etal-2023-pretraining}, 
and harmful or unsafe outputs \citep{anwar2024foundational,ye-etal-2024-toolsword,zhang-etal-2024-safetybench}. 
In such instances, it is appropriate for the model to abstain \citep{kamath-etal-2020-selective,feng-etal-2024-dont,srinivasan-etal-2024-selective} from generating unintended content.
Abstention can be employed for unanswerable \citep{sulem-etal-2022-yes,amayuelas-etal-2024-knowledge} 
or ambiguous \citep{min-etal-2020-ambigqa,kim-etal-2024-aligning} queries.
Furthermore, models may abstain when relevant parametric knowledge is absent \citep{ahdritz2024distinguishing,kim-thorne-2024-epistemology}.
Abstention can be facilitated by utilizing confidence scores of generations \citep{sun2022quantifying,kuhn2023semantic,duan-etal-2024-shifting} or
training the model for abstention capabilities \citep{zhang-etal-2024-r,sun-etal-2024-aligning,cohen2024i}.
Unlike previous approaches, this work proposes a training-free decoding method, enabling off-the-shelf models to abstain when necessary.




\begin{figure*}[t]
    \centering

    \includegraphics[width=0.99\linewidth]{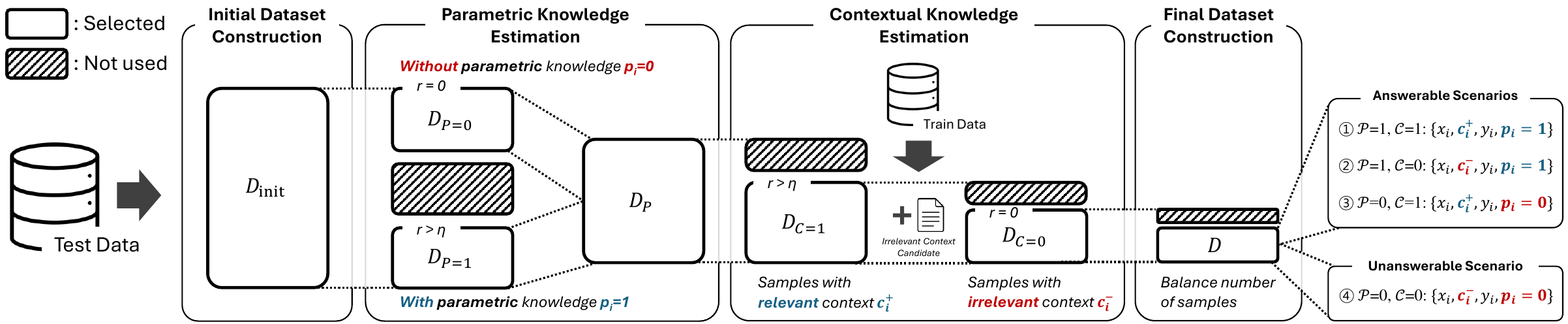}

    \caption{
        The overall process of dataset construction for the testbed.
    }
    \label{figure:data_processing_2}
\end{figure*}

\section{Testbed Design for Controlled Analysis}
\label{sec:controlled_exp}

The primary objective of this work is to enable the model to dynamically adjust its behavior based on the presence and absence of its knowledge.
Specifically, the model must effectively address all four scenarios depicted in Figure \ref{figure:intro_figure}. 
However, as we lack prior information regarding the model's possessed knowledge, it is challenging to determine and evaluate whether the model should provide an answer or abstain from a given query.
Thus, we construct a testbed by explicitly controlling the accessibility of the knowledge to simulate all the scenarios.
This section first formulates the problem and describes the setup process as illustrated in Figure \ref{figure:data_processing_2}.
Further details are in Appendix \ref{appendix:controlled_experimental_setting}.


\subsection{Problem Formulation}
\label{sec:problem_formulation}

This paper focuses on QA tasks, which facilitate a clear assessment of the knowledge usage of the model.
\textit{Parametric} knowledge (\pp{}) is defined as the knowledge the model acquires during pre-training, and \textit{contextual} knowledge (\cc{}) refers to the external knowledge provided within the input at inference time.
The knowledge is deemed \textit{relevant} if it contains information capable of generating an accurate response to the query.
For a given query $x$ and a context $c$, the objective is to produce the ground-truth answer $y$ when relevant knowledge is available or to abstain otherwise.

Figure \ref{figure:intro_figure} illustrates the scenarios addressed in this work.
Inputs are defined \textit{answerable} if one or more relevant knowledge is present (\pp{}=1 or \cc{}=1).
With relevant parametric knowledge (\pp{}=1), the model is expected to generate the correct answer regardless of $c$.  
On the other hand, the model should generate grounded on $c$ given relevant contextual knowledge (\cc{}=1).
When no relevant knowledge is available (\pp{}=0 and \cc{}=0), the query is considered \textit{unanswerable}, and the model should refuse to generate incorrect responses.
Thus, a \textit{reliable} model should properly generate an accurate answer or abstain grounded on the possessed knowledge.


\subsection{Initial Dataset Construction}
\label{sec:preprocessing}
The testbed utilizes three extractive QA datasets from the MRQA benchmark \citep{fisch-etal-2019-mrqa}:
Natural Questions (NQ) \citep{kwiatkowski-etal-2019-natural}, HotpotQA \citep{yang-etal-2018-hotpotqa}, and TriviaQA \citep{joshi-etal-2017-triviaqa}. 
Each dataset consists of a query $x_i$, an answer $y_i$, and a pre-defined context $c_i$\footnote{
    We assume the context is always factual and only focus on the relevance to the query.
    While incorporating the context's factuality is practical, we consider it orthogonal to this study.
} containing one or more answer spans.
We split $c_i$ into 100-word spans containing $y_i$ to avoid excessively long contexts.
Through preprocessing, we construct $\mathcal{D}_{\text{init}}=\{ (x_i,c_i,y_i) \}^{\mathcal{N}_{\text{init}}}_{i=1}$.


\begin{figure}[t!]
    \centering

    \begin{subfigure}{0.95\columnwidth}
    \centering
    
    \includegraphics[width=0.91\columnwidth]{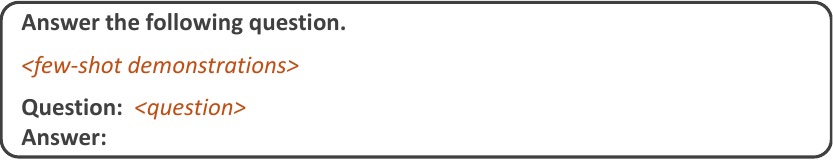}

    \caption{
        Parametric template $\mathcal{T}_p(\cdot)$.
    }
    \label{template:parametric}
    \end{subfigure}
    
    \begin{subfigure}{0.95\columnwidth}
    \centering
    \includegraphics[width=0.91\columnwidth]{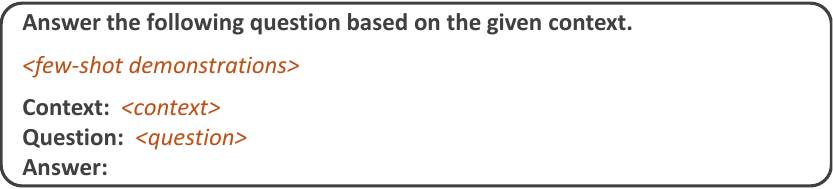}

    \caption{
        Contextual template $\mathcal{T}_c(\cdot)$.
    }
    \label{template:contextual}
    \end{subfigure}

    \begin{subfigure}{0.95\columnwidth}
    \centering
    \includegraphics[width=0.91\columnwidth]{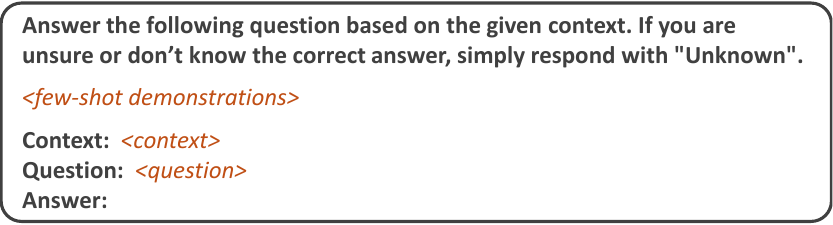}

    \caption{
        Explicit abstention template $\mathcal{T}_a(\cdot)$.
    }
    \label{template:abstention}
    \end{subfigure}

    \caption{
        List of inference templates.
    }
    \label{template:full}
\end{figure}

\subsection{Parametric Knowledge Estimation}
\label{sec:parametric_knowledge}
To estimate the model's parametric knowledge, we assess the generation consistency \citep{wang2023selfconsistency}\footnote{
    Estimating parametric knowledge is challenging \citep{shi2024detecting} due to various influencing factors.
    Since considering all potential factors is infeasible, we employ a fixed inference setting for all the experiments, which we contend best approximates the knowledge within a controlled setting.
}
for a query $x_i \in \mathcal{D}_{\text{init}}$.
We prompt the model with the parametric template $\mathcal{T}_p(x_i)$ from Figure \ref{template:parametric}, which relies solely on the model's parametric knowledge for the prediction.
By sampling $n$ responses, we compute the consistency rate $r=\frac{m}{n}$, where $m$ is the number of correct responses.
If $r=0$, we assume the model lacks relevant parametric knowledge for $x_i$.
These samples are collected as $\mathcal{D}_{\mathcal{P}=0}=\{(x_i,c_i,y_i,p_i=0)\}^{\mathcal{N}_{\mathcal{P}=0}}_{i=1}$. 
On the other hand, the model is considered to pose relevant parametric knowledge if $r>\eta$ for a pre-defined threshold $\eta$, given its consistent accuracy.
These samples are grouped into $\mathcal{D}_{\mathcal{P}=1}=\{(x_i,c_i,y_i,p_i=1)\}^{\mathcal{N}_{\mathcal{P}=1}}_{i=1}$.
The resulting dataset is defined as $\mathcal{D}_{\mathcal{P}} = \mathcal{D}_{\mathcal{P}=0} + \mathcal{D}_{\mathcal{P}=1}$.


\subsection{Contextual Knowledge Estimation}
\label{sec:contextual_knowledge}
In this stage, we select relevant and irrelevant contexts for a given query.

\paragraph{Relevant Context Selection}
We provide the model with a contextual template $\mathcal{T}_c(c_i,x_i)$ from Figure \ref{template:contextual}, where $(x_i, c_i) \in \mathcal{D}_{\mathcal{P}}$.
By providing $c_i$ as the input, the model can leverage contextual knowledge.
We further verify the relevance of $c_i$ by computing the consistency rate $r$.
Samples with $r>\eta$ are defined as the relevant context $c_i^{+}$ and are grouped into $\mathcal{D}_{\mathcal{C}=1}=\{(x_i,c_i^{+},y_i,p_i)\}^{\mathcal{N}_{\mathcal{C}=1}}_{i=1}$.

\paragraph{Irrelevant Context Selection}
For each sample $(x_i,c_i^{+})\in\mathcal{D}_{\mathcal{C}=1}$, we select an irrelevant context candidate $c^{\text{train}}_{j}$ from the training set.
The candidate is selected with the highest SBERT \citep{reimers-gurevych-2019-sentence} embedding similarity to $c_i^{+}$ to avoid overly unrelated contexts.
We then prompt the model with $\mathcal{T}_c(c^{\text{train}}_{j},x_i)$ and measure the consistency rate $r$.
Only candidates with $r=0$ are considered irrelevant context $c^{-}_i$, ensuring that $c^{-}_i$ does not provide any unintended relevant information.
The resulting dataset is defined as $\mathcal{D}_{\mathcal{C}=0}=\{(x_i,c_i^{+},c_i^{-},y_i,p_i)\}^{\mathcal{N}_{\mathcal{C}=0}}_{i=1}$.


\subsection{Final Dataset Construction}
Finally, we randomly select an equal number of samples with $p_i=0$ and $p_i=1$, constructing $\mathcal{D}=\{(x_i,c_i^{+},c_i^{-},y_i,p_i)\}^{\mathcal{N}}_{i=1}$. 
Note that the number of selected data varies across models due to their distinct knowledge boundaries.






\section{Contrastive Decoding with Abstention}
\label{sec:method}
Contrastive Decoding with Abstention (\method{}) is a novel decoding method integrating abstention within the \cd{} process.
This section provides a detailed description of the overall process.

\subsection{Preliminary}
Given a model $\theta$ at decoding step $t$, the parametric knowledge distribution $d^p_t$ and the contextual knowledge distribution $d^c_t$ is defined as follows:
\begin{equation}
    \begin{split}
    &d^p_t = \text{logit}_{\theta}(y_t\ |\ \mathcal{T}_p(x,y_{<t})), \\    
    &d^c_t = \text{logit}_{\theta}(y_t\ |\ \mathcal{T}_c(c,x,y_{<t}))
    \label{eq:distribution}.
    \end{split}
\end{equation}
\cd{} measures the final output distribution $d^o_t$ as an ensemble of $d^p_t$ and $d^c_t$ as:
\begin{equation}
    d^o_t = d^p_t + w^c_t\ (d^c_t - d^p_t)
    \label{eq:org_cd}
\end{equation}
The weight $w^c_t$ is intended to quantify the relevance of $c$, ensuring a higher weight when $c$ is relevant.

\subsection{Incorporating Abstention}
To enable \method{} to properly abstain,
we incorporate the abstention distribution $d^a_t$ computed from an explicit abstention instruction $\mathcal{T}_a(\cdot)$ in Figure \ref{template:abstention}.
\begin{equation}
    d^a_t = \text{logit}_{\theta}(y_t\ |\ \mathcal{T}_a(c,x,y_{<t}))
    \label{eq:abstain_distribution}
\end{equation}
\method{} expands Eq. \ref{eq:org_cd} by applying $d^a_t$ for the final output distribution, where the weight for $d^a_t$ is defined as $w^a_t=1-w^p_t-w^c_t$.
\begin{equation}
\begin{aligned}
    d^o_t &= d^p_t + w^c_t\ (d^c_t - d^p_t) + w^a_t\ (d^a_t - d^p_t) \\
    &= (1-w^c_t-w^a_t)\ d^p_t + w^c_t\ d^c_t + w^a_t\ d^a_t \\
    &= w^p_t\ d^p_t + w^c_t\ d^c_t + (1-w^p_t-w^c_t)\ d^a_t
    \label{eq:ours}
\end{aligned}
\end{equation}
Intuitively, $w^a_t$ decreases when the model is confident of a possessed knowledge, whereas it increases when both knowledge are uncertain.

\subsection{Knowledge Relevance Assessment via Uncertainty Calibration}
A key requirement for \method{} is that the weights $w^p_t$ and $w^c_t$ should effectively quantify the relevance of the corresponding knowledge.
We assess the relevance as the uncertainty of the corresponding knowledge regarding $x$.
Specifically, we utilize the entropy \citep{malinin2021uncertainty,ABDAR2021243}, 
a widely used measure to assess the uncertainty of the knowledge for a query \citep{kuhn2023semantic,duan-etal-2024-shifting},
particularly prevalent in \cd{} \citep{kim-etal-2024-adaptive,qiu2024entropy}.
For an output distribution $d$, the entropy $\text{\entropy}$ is defined as:
\begin{equation}
    \text{\entropy} = -\sum_{i=1}^{|\mathcal{V}|}d_i\ \text{log}\ d_i,
    \label{eq:entropy}
\end{equation}
where $d_i$ it the $i^{\text{th}}$ token of the vocabulary $\mathcal{V}$.
The parametric uncertainty $\text{\entropy}^p_t$ and contextual uncertainty $\text{\entropy}^c_t$ are derived from their respective distributions $d^p_t$ and $d^c_t$.

Nonetheless, directly comparing $\text{\entropy}^p_t$ and $\text{\entropy}^c_t$ are imprecise since they are conditioned on distinct inputs and possibly miscalibrated.
To address this, we ``calibrate'' \citep{Zhao2021CalibrateBU,holtzman-etal-2021-surface,he-etal-2024-prompt} the uncertainty measures by accounting for the model's intrinsic bias.
Specifically, we estimate the bias with a ``content-free'' \nullprompt{} prompt, 
replacing specific inputs $x_i$ and $c_i$ with placeholder tokens $\Bar{x}$ and $\Bar{c}$ to remove any specific semantic information. 
Applying the templates $\mathcal{T}_p(\Bar{x})$ and $\mathcal{T}_c(\Bar{c},\Bar{x})$ yields the parametric \nullprompt{} distribution $\Bar{d}^p_t$ and the contextual \nullprompt{} distribution $\Bar{d}^c_t$, along with their corresponding entropy values $\Bar{\text{\entropy}}^p_t$ and $\Bar{\text{\entropy}}^c_t$.
The confidence for the knowledge is quantified as the additional information provided by the input relative to the \nullprompt{} prompt.
\begin{equation}
    \begin{split}
    r^p_t = \tfrac{\text{max}(\text{\entropy}^p_t-\Bar{\text{\entropy}}^p_t,\ 0)}{\Bar{\text{\entropy}}^p_t}, \ r^c_t = \tfrac{\text{max}(\text{\entropy}^c_t-\Bar{\text{\entropy}}^c_t,\ 0)}{\Bar{\text{\entropy}}^c_t}.
    \end{split}
\end{equation}
We obtain the final weights $w^p_t$ and $w^c_t$ from Eq. \ref{eq:ours} by normalizing $r^p_t$ and $r^c_t$, respectively.
\begin{equation}
    w^p_t = \tfrac{r^p_t}{r^p_t+r^c_t}\ r^p_t, \ w^c_t = \tfrac{r^c_t}{r^p_t+r^c_t}\ r^c_t 
\end{equation}
When a particular knowledge source provides substantial additional information relative to the \nullprompt{} prompt, it is assigned a higher weight, whereas the absence of knowledge results in a reduced weight.

\subsection{\method{} with Momentum (\methodm{})}
At each decoding step, previous content may unintentionally steer the model toward irrelevant knowledge. 
To mitigate this, we apply momentum, updating the current weight $w_t$ as a convex combination with the previous weight $w_{t-1}$.
\begin{equation}
    w_t \leftarrow \alpha\ w_{t-1} + (1-\alpha)\ w_t
\end{equation}
Here, the hyperparameter $\alpha$ controls the influence of the previous step on the current step.
Applying momentum helps stabilize the decoding process by smoothing abrupt weight changes.




\begin{figure}[t]
    \begin{center}
        \includegraphics[width=0.82\columnwidth]{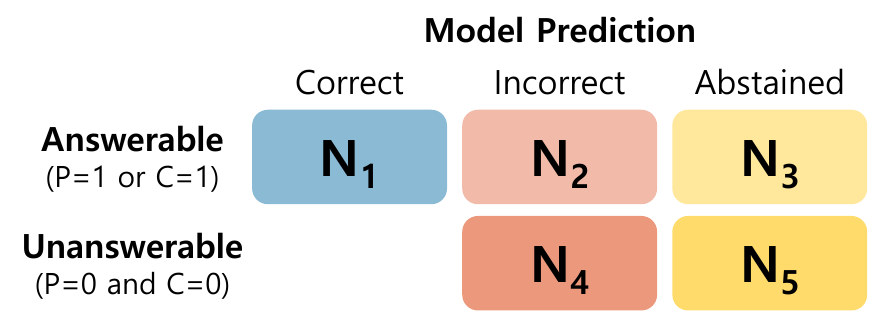}
          \caption{
            Illustration of all possible results. 
            The model should generate correct answers for answerable queries (\1) and abstain for unanswerable queries (\5). 
            Any other responses (\2, \3, \4) are classified as incorrect.
          }
          \label{fig:metric}
    \end{center}
\end{figure}

\definecolor{my_gray}{HTML}{E8E8E8}
\newcolumntype{g}{>{\columncolor{my_gray}}c}

\begin{table*}[t!]
    \centering    
    \resizebox{0.97\textwidth}{!}{

  \begin{tabular}{cgggggcccccgggggccccc}
\toprule

Backbone &
\multicolumn{5}{c}{\begin{tabular}[c]{@{}c@{}} {\scshape Llama3 8B}\\ {\scshape Instruct}\end{tabular}} &
\multicolumn{5}{c}{\begin{tabular}[c]{@{}c@{}} {\scshape Llama2 7B}\\ {\scshape Chat}\end{tabular}} &
\multicolumn{5}{c}{\begin{tabular}[c]{@{}c@{}} {\scshape Llama2 13B}\\ {\scshape Chat}\end{tabular}} &
\multicolumn{5}{c}{\begin{tabular}[c]{@{}c@{}} {\scshape Mistral 7B}\\ {\scshape Instruct}\end{tabular}} \\
\midrule
\rowcolor{white}
Method & \ans{} & \abs{} & RS & Acc. & Cov. &
\ans{} & \abs{} & RS & Acc. & Cov. &
\ans{} & \abs{} & RS & Acc. & Cov. &
\ans{} & \abs{} & RS & Acc. & Cov. \\ 
\midrule
\multicolumn{21}{c}{\textit{NQ}} \\ 
\midrule
\context{} &
57.26 & 2.18 & 50.22 & 49.95 & 50.23 & 
57.13 & 0.09 & 49.98 & 49.97 & 49.98 & 
57.14 & 0.37 & 50.00 & 49.95 & 50.00 & 
57.04 & 0.25 & 49.88 & 49.85 & 49.88 \\
\cad{} & 
55.34 & 2.14 & 48.56 & 48.29 & 48.56 & 
54.78 & 0.17 & 47.94 & 47.92 & 47.94 & 
45.57 & 0.48 & 39.90 & 39.84 & 39.90 & 
55.03 & 1.15 & 48.21 & 48.07 & 48.21 \\
\acd{} & 
71.36 & 0.92 & 62.50 & \textbf{62.39} & 62.50 & 
64.55 & 0.15 & 56.48 & \textbf{56.46} & 56.48 & 
66.66 & 0.18 & 58.33 & \textbf{58.31} & 58.33 & 
68.48 & 0.00 & 59.93 & \textbf{59.93} & 59.93 \\
\abstain{} & 
60.22 & 52.05 & 48.17 & 38.36 & 57.27 & 
38.73 & 45.16 & 26.73 & 20.54 & 42.11 & 
50.49 & 46.62 & 37.83 & 29.70 & 48.93 & 
61.86 & 53.29 & 52.85 & 42.83 & 59.70 \\
\sa{} & 
57.23 & 48.11 & 43.91 & 35.06 & 53.70 & 
56.81 & 10.95 & 50.17 & 48.66 & 50.23 & 
59.37 & 20.80 & 52.98 & 49.89 & 53.21 & 
62.52 & 49.03 & 57.49 & 48.76 & 59.54 \\
\ent{} & 
64.06 & 53.34 & 55.53 & 45.15 & 60.90 & 
58.00 & 41.09 & 51.98 & 44.47 & 54.08 & 
59.23 & 42.61 & 52.75 & 44.85 & 55.19 & 
61.91 & 56.29 & 55.48 & 44.63 & 60.33 \\
\acda{} & 
63.56 & 52.46 & 53.99 & 43.82 & 60.11 & 
48.82 & 39.52 & 37.55 & 30.41 & 45.66 & 
57.96 & 46.34 & 49.28 & 40.27 & 54.43 & 
61.46 & 54.02 & 55.80 & 45.58 & 59.56 \\
\fsb{} & 
69.27 & 54.94 & 59.64 & 49.02 & 65.09 & 
55.04 & 47.26 & 43.26 & 34.47 & 52.20 & 
62.44 & \underline{47.60} & 53.66 & 44.46 & 58.19 & 
66.71 & 55.51 & 58.95 & 48.32 & 63.65 \\
\midrule
\method{} & 
\underline{72.06} & \textbf{55.49} & \underline{62.95} & 52.28 & \underline{67.51} & 
\underline{66.86} & \underline{47.52} & \underline{59.86} & 51.22 & \underline{62.38} & 
\underline{68.62} & 47.14 & \underline{61.63} & 53.16 & \underline{63.76} & 
\underline{69.68} & \underline{56.47} & \underline{61.45} & 50.61 & \underline{66.09} \\
\methodm{} & 
\textbf{73.15} & \underline{55.47} & \textbf{63.72} & \underline{53.16} & \textbf{68.30} & 
\textbf{69.99} & \textbf{47.60} & \textbf{62.28} & \underline{53.62} & \textbf{64.81} & 
\textbf{70.66} & \textbf{48.12} & \textbf{63.18} & \underline{54.46} & \textbf{65.48} & 
\textbf{71.00} & \textbf{56.46} & \textbf{62.30} & \underline{51.45} & \textbf{67.03} \\ 
 \midrule
\multicolumn{21}{c}{\textit{HotpotQA}} \\ 
\midrule
\context{} & 
57.15 & 1.19 & 50.08 & 49.93 & 50.08 & 
57.14 & 0.00 & 49.98 & 49.98 & 49.98 & 
57.16 & 0.10 & 50.00 & 49.98 & 50.00 & 
56.99 & 0.09 & 49.88 & 49.87 & 49.88 \\
\cad{} & 
55.78 & 1.88 & 48.89 & 51.99 & 48.89 & 
54.68 & 0.04 & 47.84 & 47.84 & 47.84 & 
54.60 & 0.07 & 47.76 & 47.75 & 47.76 & 
53.74 & 0.20 & 47.04 & 47.01 & 47.04 \\
\acd{} & 
74.36 & 0.57 & 65.09 & \textbf{65.02} & 65.09 & 
69.27 & 0.00 & 60.60 & \underline{60.60} & 60.60 & 
69.80 & 0.07 & 61.15 & \textbf{61.14} & 61.15 & 
72.72 & 0.09 & 63.64 & \textbf{63.62} & 63.64 \\
\abstain{} & 
66.88 & 56.58 & 56.23 & 45.21 & 63.55 & 
47.29 & 48.65 & 33.49 & 26.18 & 48.18 & 
57.17 & 51.95 & 43.02 & 33.89 & 55.10 & 
61.12 & 54.66 & 53.06 & 42.55 & 59.23 \\
\sa{} & 
50.64 & 48.63 & 33.86 & 26.80 & 49.70 & 
58.58 & 17.35 & 52.19 & 49.70 & 52.33 & 
58.22 & 14.36 & 51.87 & 49.89 & 51.95 & 
61.33 & 43.36 & 56.36 & 48.95 & 57.70 \\
\ent{} & 
67.08 & 56.44 & 57.92 & 46.94 & 63.88 & 
59.00 & 45.11 & 52.64 & 44.16 & 55.51 & 
59.24 & 45.47 & 53.01 & 44.49 & 55.81 & 
63.31 & \textbf{60.07} & 57.19 & 45.65 & 62.40 \\
\acda{} & 
65.88 & 54.67 & 57.41 & 46.80 & 62.63 & 
57.79 & \underline{51.11} & 47.90 & 38.19 & 55.61 & 
62.00 & 52.13 & 54.03 & 43.95 & 59.16 & 
61.02 & 50.93 & 57.01 & 48.10 & 58.95 \\
\fsb{} & 
74.89 & 58.51 & 66.21 & 55.05 & 70.55 & 
65.63 & \textbf{53.82} & 55.73 & 45.37 & 62.04 & 
68.68 & 54.16 & 60.41 & 50.05 & 64.79 & 
74.32 & 52.18 & 68.20 & 59.27 & 69.94 \\
\midrule
\method{} & 
\underline{78.71} & \underline{62.50} & \underline{70.20} & 58.36 & \underline{74.52} & 
\underline{73.39} & 42.41 & \underline{66.96} & 60.15 & \underline{67.82} & 
\textbf{73.69} & \underline{56.81} & \underline{68.98} & \underline{59.53} & \textbf{70.44} & 
\underline{76.28} & 55.84 & \underline{69.43} & 59.50 & \underline{71.83} \\
\methodm{} & 
\textbf{79.32} & \textbf{62.59} & \textbf{70.64} & \underline{58.78} & \textbf{74.99} & 
\textbf{74.09} & 42.31 & \textbf{67.50} & \textbf{60.70} & \textbf{68.37} & 
\underline{73.66} & \textbf{56.89} & \textbf{68.92} & 59.42 & \underline{70.42} & 
\textbf{76.94} & \underline{56.67} & \textbf{69.98} & \underline{59.85} & \textbf{72.49} \\ 
 \midrule
\multicolumn{21}{c}{\textit{TriviaQA}} \\ 
\midrule
\context{} & 
57.29 & 2.24 & 50.23 & 49.95 & 50.23 & 
57.17 & 0.26 & 50.02 & 49.99 & 50.02 & 
57.15 & 0.47 & 50.03 & 49.97 & 50.03 & 
57.13 & 0.35 & 50.00 & 49.96 & 50.00 \\
\cad{} & 
55.83 & 0.65 & 48.87 & 48.79 & 48.87 & 
55.11 & 0.25 & 48.22 & 48.19 & 48.22 & 
55.81 & 0.32 & 48.84 & 48.80 & 48.84 & 
54.71 & 0.57 & 47.89 & 47.82 & 47.89 \\
\acd{} & 
76.79 & 3.09 & 67.38 & \textbf{66.99} & 67.39 & 
72.86 & 0.18 & 63.74 & \textbf{63.72} & 63.74 & 
\underline{72.49} & 0.42 & 63.43 & \textbf{63.38} & 63.43 & 
75.01 & 0.05 & 65.62 & \textbf{65.62} & 65.62 \\
\abstain{} & 
67.46 & 57.31 & 56.21 & 45.07 & 64.10 & 
59.19 & 50.45 & 47.04 & 37.87 & 56.27 & 
48.53 & 46.74 & 35.11 & 27.52 & 47.88 & 
60.73 & 53.16 & 51.25 & 41.22 & 58.42 \\
\sa{} & 
52.40 & 48.38 & 36.73 & 29.01 & 50.61 & 
57.93 & 8.97 & 51.05 & 49.84 & 51.09 & 
58.09 & 12.07 & 51.53 & 49.85 & 51.59 & 
62.18 & 48.60 & 57.19 & 48.55 & 59.20 \\
\ent{} & 
66.17 & 57.26 & 56.50 & 45.33 & 63.35 & 
60.21 & 47.84 & 53.99 & 45.00 & 57.10 & 
60.21 & 48.31 & 54.31 & 45.32 & 57.27 & 
62.21 & \underline{56.59} & 56.23 & 45.48 & 60.71 \\
\acda{} & 
66.67 & 56.73 & 58.81 & 57.88 & 63.87 & 
61.21 & 50.52 & 53.09 & 43.46 & 58.18 & 
58.42 & 49.72 & 49.22 & 39.54 & 55.61 & 
61.62 & 51.62 & 56.81 & 47.48 & 59.48 \\
\fsb{} & 
77.02 & 59.84 & 68.55 & 57.24 & 72.62 & 
69.50 & \underline{51.88} & 60.41 & 50.59 & 64.78 & 
66.21 & \underline{52.08} & 56.12 & 45.98 & 61.91 & 
\textbf{77.67} & 47.53 & \textbf{70.69} & 62.72 & \underline{72.07} \\
\midrule
\method{} & 
\underline{80.39} & \textbf{65.67} & \underline{72.35} & 60.01 & \underline{76.67} & 
\textbf{73.70} & 51.29 & \textbf{67.29} & \underline{58.43} & \textbf{69.06} & 
71.08 & 51.44 & \underline{64.08} & 54.78 & \underline{66.55} & 
75.76 & 56.35 & 67.57 & 57.21 & 71.21 \\
\methodm{} & 
\textbf{80.93} & \underline{65.66} & \textbf{72.74} & \underline{60.40} & \textbf{77.07} & 
\underline{73.47} & \textbf{52.10} & \underline{67.11} & 58.04 & \underline{69.00} & 
\textbf{73.12} & \textbf{53.46} & \textbf{65.82} & \underline{56.09} & \textbf{68.52} & 
\underline{76.95} & \textbf{57.06} & \underline{68.38} & \underline{57.84} & \textbf{72.21} \\ 
 \bottomrule
\end{tabular}

    }
    \caption{
        Experimental results on three different datasets.
        For each dataset, the \textbf{best method} is highlighted in bold, and the \underline{second-best method} is underlined.
        \both{} outperforms all the baselines across different metrics.
    }
    \label{table:main_results}
\end{table*}

\section{Experiments}

\subsection{Experimental Setting}
The experiments utilize the testbed from Section \ref{sec:controlled_exp} and four instruction-tuned models including \ml{} \citep{dubey2024llama}, {\scshape Llama2 7B \& 13B Chat} \citep{touvron2023llama}, and \mm{} \citep{jiang2023mistral}.
Results are averaged over three different random seeds.
Further details are stated in Appendix \ref{appendix:experiments}.

\subsection{Evaluation Metric}
To measure the overall performance across two distinct tasks, we adopt three metrics, each reflecting unique aspects of performance, based on the five possible results in Figure \ref{fig:metric}.

\paragraph{Answerable Prediction F1 (\ans{})} 
For answerable queries, we compute \ans{} \citep{kim-etal-2024-aligning} as the harmonic mean of 
precision ($\frac{\text{\1}}{\text{\1}+\text{\2}+\text{\4}}$) and
recall ($\frac{\text{\1}}{\text{\1}+\text{\2}+\text{\3}}$).
The prediction is considered correct if it contains the ground-truth answer \citep{mallen-etal-2023-trust,NEURIPS2023_d842425e}.

\paragraph{Abstention F1 (\abs{})} 
The model should abstain from incorrect responses for unanswerable queries while minimizing over-abstention.
\abs{} \citep{kim-etal-2024-aligning} measures such behaviors by incorporating both precision ($\frac{\text{\5}}{\text{\3}+\text{\5}}$) and recall ($\frac{\text{\5}}{\text{\4}+\text{\5}}$).
A prediction is deemed as an abstention if it contains any pre-defined abstention phrases \citep{amayuelas-etal-2024-knowledge,kim-etal-2024-aligning}.

\paragraph{Reliability Score (RS)} 
RS \citep{xu2024rejection} is the weighted sum of accuracy (Acc., $\frac{\text{\1}}{\text{\full}}$) and coverage (Cov., $\frac{\text{\1}+\text{\3}+\text{\5}}{\text{\full}}$), 
where \full{} is the total number of samples and $\alpha$ is set as the answer rate ($1-\frac{\text{\3}+\text{\5}}{\text{\full}}$).
\begin{equation}
    \text{RS}(\alpha) = \alpha \times \text{Cov.} + (1-\alpha) \times \text{Acc.}
\end{equation}
RS prioritizes accuracy at a lower answer rate while avoiding errors with a higher coverage at a high answer rate.
We also report the accuracy and coverage for a more thorough analysis.

\subsection{Baselines}
To evaluate the effectiveness of our approach, we compare different inference methods as baselines.

\paragraph{Direct Prompting}
includes \textbf{contextual prompting (\context{})} employing $\mathcal{T}_c(\cdot)$ and 
\textbf{abstention prompting (\abstain{})} utilizing $\mathcal{T}_a(\cdot)$, with an explicit instruction for abstention.

\paragraph{\sa{}} prompts the model with $\mathcal{T}_c(\cdot)$ and further verifies the generation \citep{kadavath2022language}.
Predictions verified as ``unknown'' are abstained.

\paragraph{Context-aware Decoding (\cad{})}
amplifies contextual influence by gauging $d^o_t = d^c_t + w^c_t\ (d^c_t - d^p_t)$ with a fixed $w^c_t$ during decoding \citep{shi-etal-2024-trusting}.

\paragraph{\ent{}} measures the entropy (Eq. \ref{eq:entropy}) of the generated tokens when prompted with $\mathcal{T}_c(\cdot)$ \citep{10820047}.
We measure four different variants --- first-token, average, maximum, and minimum entropy --- and report the first-token entropy with the best performance.
If the measure exceeds a pre-defined threshold, the prediction is deemed uncertain and thus abstained.

\paragraph{Adaptive Contrastive Decoding (\acd{})} 
follows Eq. \ref{eq:org_cd} where $w^c_t=1-\tfrac{\text{\entropy}^c_t}{\text{\entropy}^p_t+\text{\entropy}^c_t}$ 
 \citep{kim-etal-2024-adaptive}.

\paragraph{ACD with Abstention (\acda{})}
expands \acd{} to perform abstention where $w^c_t = 1 - \tfrac{\text{\entropy}^c_t}{\text{\entropy}^p_t+\text{\entropy}^c_t+\text{\entropy}^a_t}$ and $w^a_t = 1 - \tfrac{\text{\entropy}^a_t}{\text{\entropy}^p_t+\text{\entropy}^c_t+\text{\entropy}^a_t}$ following Eq. \ref{eq:ours}.

\paragraph{First Step Branching (\fsb{})}
compares the first-token entropy $\text{\entropy}^p_1$, $\text{\entropy}^c_1$, and $\text{\entropy}^a_1$ when prompted with $\mathcal{T}_p(\cdot)$, $\mathcal{T}_c(\cdot)$, and $\mathcal{T}_a(\cdot)$, respectively. 
We select the most certain method and continue the generation with the selected method.


\subsection{Main Results}
The main results are presented in Table \ref{table:main_results}.
\paragraph{Methods not accounting for abstention fail to handle unanswerable queries.}
Method such as \context{}, \cad{}, and \acd{} exhibit near-zero \abs{}, indicating their inability to handle unanswerable queries.
Moreover, the negligible gap between their accuracy and coverage further confirms their incapability to abstain.

\paragraph{Incorporating abstention enhances the handling of unanswerable queries.}
\abstain{} and \sa{} exhibit biased abstentions, resulting in low accuracy and high coverage.
\acda{} and \ent{} perform abstention to some extent, but they struggle to balance between accurate generation and abstention.
\fsb{} emerges as the strongest among the baseline, effectively addressing (un)answerable queries.
Overall, incorporating abstention does provide the model of the ability to abstain, but they fail to effectively balance the trade-off between accurate generation and appropriate abstention.

\paragraph{\both{} exhibits superior performance across all datasets.} 

\both{} outperforms all the baselines on \abs{} and RS, 
properly abstaining unanswerable queries.
Moreover, its effective handling of answerable queries exhibits the highest \ans{} and the second-best accuracy following \acd{}.



\section{Ablation Study}
\label{sec:ablations}
This section presents ablation studies of \both{}. 
Unless otherwise specified, all experiments are conducted on \ml{} with the testbed from Section \ref{sec:controlled_exp}.
Methods capable of abstention, including \abstain{}, \sa{}, \ent{}, \fsb{}, and \acda{}, are utilized for comparison.
Further details and results are in Appendix \ref{appendix:ablation}.

\definecolor{correct}{RGB}{170, 199, 218}
\definecolor{abstained}{RGB}{181, 184, 175}

\begin{figure}[t]
    \centering
    
    \includegraphics[width=0.99\columnwidth]{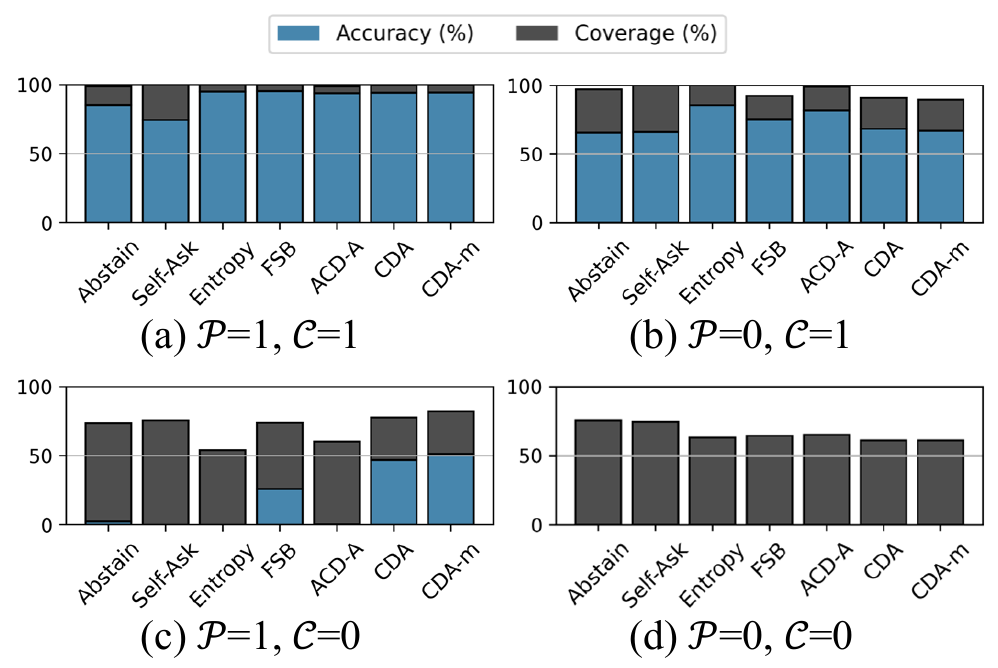}

    \caption{
        The \colorbox{correct}{accuracy} and \colorbox{abstained}{coverage} for all the scenarios regarding both knowledge.
        \both{} effectively balances between correct predictions and abstentions, especially in the presence of irrelevant contexts ($\mathcal{C}$=0).
    }
    \label{figure:scenario_2}
\end{figure}

\subsection{Analysis of Different Scenarios}
Figure \ref{figure:scenario_2} depicts the accuracy (in blue) and coverage (in gray) for each scenario in the NQ dataset.
Note that the main objective is to balance between accurate generation when relevant knowledge is present (\pp{}=1 or \cc{}=1) and abstention when such knowledge is absent (\pp{}=0 and \cc{}=0).
Most baselines exhibit over-abstention, particularly with irrelevant context.
These methods fail to utilize relevant parametric knowledge, leading either to biased abstention or incorrect generations.
In contrast, \both{} robustly leverages proper knowledge while maintaining balanced abstention capability.

\subsection{Ablation on Momentum Weight}
\label{sec:momentum_ablation}
\begin{figure}[t]
    \begin{center}

    \includegraphics[width=0.98\columnwidth]{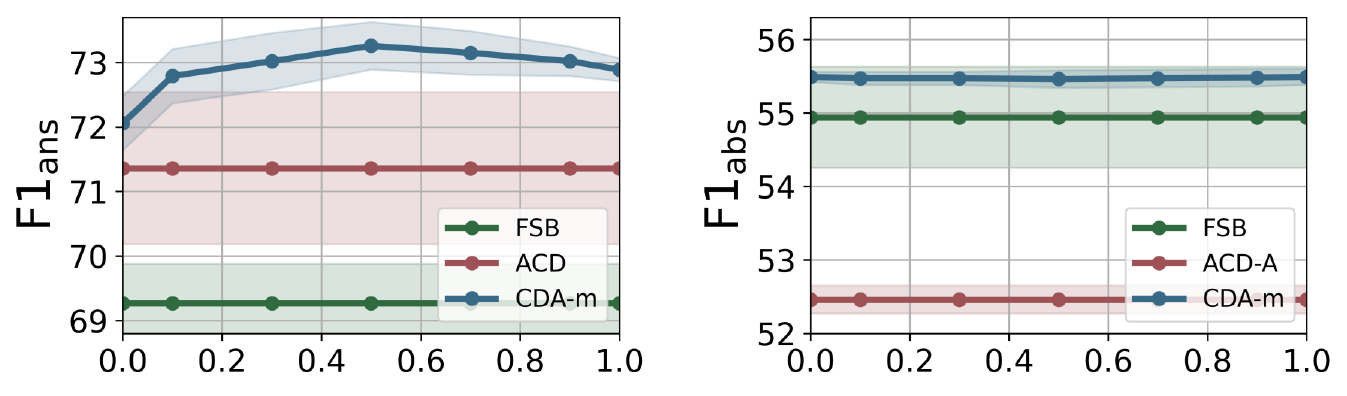}
        
          \caption{
            \ans{} and \abs{} according to different $\alpha$ values.
            Applying momentum significantly improves \ans{}.
          }
          \label{figure:alpha_ablation}
    \end{center}
\end{figure}

In this section, we investigate the impact of momentum weight $\alpha$ on \methodm{}.
Figure \ref{figure:alpha_ablation} displays \ans{} and \abs{} of $\alpha$ from 0.0 to 1.0, along with the best-performing baselines in the NQ dataset. 
We can observe that \abs{} remains stable while \ans{} improves when applying momentum.
As intended, momentum reduces incorrect generations while preserving abstention capabilities.
Overall, \methodm{} outperforms the strongest baselines regardless of $\alpha$.
A further case study is conducted in Appendix \ref{appendix:momentum}.

\subsection{Effect of Calibration}
\both{} leverages calibrated uncertainty measures to quantify the relevance of different knowledge.
To evaluate the effect of calibration, we modify Eq. \ref{eq:ours} to employ non-calibrated measure by directly setting $r^p_t=\text{\entropy}^p_t$ and $r^c_t=\text{\entropy}^c_t$. 
As shown in Table \ref{table:calibration_ablation}, this change significantly degrades the performance across all metrics by up to 14 points.
Directly utilizing uncalibrated measures yields suboptimal results, thus highlighting the importance of the additional calibration step in \both{}.


\begin{table}[t]
    \centering    
    \resizebox{0.90\linewidth}{!}{

        \begin{tabular}{clccccc}
        \toprule
        Dataset & \multicolumn{1}{c}{Method} & \ans{} & \abs{} & RS & Acc. & Cov. \\ 
        \midrule
        \multirow{2}{*}{NQ} & CDA & \textbf{72.06} & \textbf{55.49} & \textbf{62.95} & \textbf{52.28} & \textbf{67.51} \\
         & \multicolumn{1}{r}{w/o calibration} & 59.35 & 52.06 & 47.89 & 38.04 & 56.79 \\ 
         \midrule
        \multirow{2}{*}{HotpotQA} & CDA & \textbf{78.71} & \textbf{62.50} & \textbf{70.20} & \textbf{58.36} & \textbf{74.52} \\
         & \multicolumn{1}{r}{w/o calibration} & 66.94 & 56.39 & 56.42 & 45.43 & 63.55 \\ 
         \midrule
        \multirow{2}{*}{TriviaQA} & CDA & \textbf{80.39} & \textbf{65.66} & \textbf{72.35} & \textbf{60.01} & \textbf{76.67} \\
         & \multicolumn{1}{r}{w/o calibration} & 67.56 & 57.24 & 56.46 & 45.32 & 64.17 \\ 
         \bottomrule
        \end{tabular}
    
    }
    \caption{
        The effects of applying calibration. 
        We can observe significant degradation without calibration.
    }
    \label{table:calibration_ablation}
\end{table}

\subsection{Comparison with Training-based Methods}
Training models to abstain have demonstrated notable performance. 
Hence, we compare \methodm{} with instruction-tuning \citep{NEURIPS2022_b1efde53,yang2024alignment} the model to perform abstention.

\paragraph{Experimental Setting}
The training data are labeled according to parametric and contextual knowledge.
Following the procedure described in Section \ref{sec:controlled_exp}, we verify whether the model possesses relevant knowledge for each training sample.
Samples with relevant knowledge are labeled with the ground-truth answer $y$, while samples without any relevant knowledge are labeled with a pre-defined abstention response $y_{\text{abs}}$ (e.g., ``unknown'').
The model is then trained to generate the label given $\mathcal{T}_c(c, x)$. 
For evaluation, we utilize the instruction-tuned model to generate an output given $\mathcal{T}_c(c, x)$.

\paragraph{Experimental Results}
Table \ref{table:trained_methods} presents the in-domain (IND) and out-of-domain (OOD) results of instruction-tuning.
\methodm{} consistently outperforms instruction-tuning, even in IND scenarios. 
Furthermore, while training often tailors the model to specific domains, resulting in significant performance drops in OOD settings, \methodm{} demonstrates superior generalization capabilities.
This robustness makes \both{} a more applicable solution for practical scenarios.


\definecolor{bg}{HTML}{b1d4e0}

\begin{table}[t]
    \centering
    
    \resizebox{0.95\linewidth}{!}{
        \begin{tabular}{cccccccccc}
        \toprule
        Target ($\rightarrow$) & \multicolumn{3}{c}{NQ} & \multicolumn{3}{c}{HotpotQA} & \multicolumn{3}{c}{TriviaQA} \\ 
        \midrule
        Source ($\downarrow$) & \ans{} & \abs{} & RS & \ans{} & \abs{} & RS & \ans{} & \abs{} & RS \\ 
        \midrule
        NQ & 
        \cellcolor{bg}\underline{66.37} & \cellcolor{bg}43.87 & \cellcolor{bg}\underline{61.12} & 
        64.43 & 48.79 & 59.65 & 
        67.31 & 49.00 & \underline{62.14} \\
        HotpotQA & 
        64.86 & 42.73 & 59.80 & 
        \cellcolor{bg}\underline{74.06} & \cellcolor{bg}\underline{57.57} & \cellcolor{bg}\underline{68.76} & 
        65.26 & 50.69 & 61.17 \\
        TriviaQA & 
        65.15 & \underline{46.37} & 57.70 & 
        66.68 & 49.82 & 60.58 & 
        \cellcolor{bg}\underline{67.60} & \cellcolor{bg}\underline{53.68} & \cellcolor{bg}61.84 \\ 
        \midrule
        \methodm{} & 
        \textbf{73.15} & \textbf{55.47} & \textbf{63.72} & 
        \textbf{79.32} & \textbf{62.59} & \textbf{70.64} & 
        \textbf{80.93} & \textbf{65.66} & \textbf{72.74} \\ 
        \bottomrule
        \end{tabular}
        
    }
    \caption{
        The results of instruction-tuning. 
        \colorbox{bg}{In-domain} results, the \textbf{best results}, and the \underline{second-based results} are highlighted.
        \methodm{} displays superior performance across all the scenarios. 
    }
    \label{table:trained_methods}
\end{table}

\begin{figure}[t]
    \begin{center}
        \includegraphics[width=0.85\columnwidth]{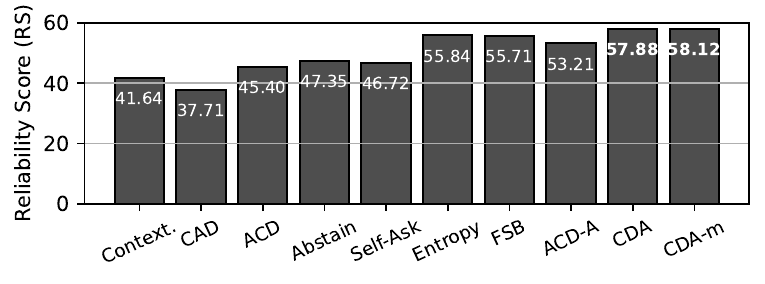}
          \caption{
            Average Reliability Score (RS) in RAG settings.
            \both{} outperforms all the baselines.
          }
          \label{figure:rag_results}
    \end{center}
\end{figure}

\subsection{Evaluation on RAG Setting}
To evaluate \method{} in practical, real-world scenarios, we conduct experiments within the RAG setting.

\paragraph{Experimental Setting}
We utilize {\scshape Contriever-msmarco} \citep{izacard2022unsupervised} as a retriever, 
and the top-1 context is retrieved from the Wikipedia contexts.\footnote{Wikipedia dump from Dec. 2018.}
Unlike the controlled setting, where the presence of both knowledge is precisely estimated, 
the prior knowledge of the given query is unknown in the RAG setting.
Since it is difficult to determine the answerablility of the given query, we evaluate solely based on the Reliability Score (RS).

\paragraph{Experimental Results}
Figure \ref{figure:rag_results} displays the average results in the RAG setting.
Similar to the main experiments, methods without abstention capabilities demonstrate poor performance,
while \fsb{} and \ent{} emerge as strong baselines.
Overall, \both{} outperform all the baselines, highlighting the effectiveness in the practical RAG setting.

\begin{figure}[t]
    \centering
    \begin{subfigure}{0.95\columnwidth}
        
        \resizebox{\linewidth}{!}{
            \includegraphics[width=0.94\linewidth]{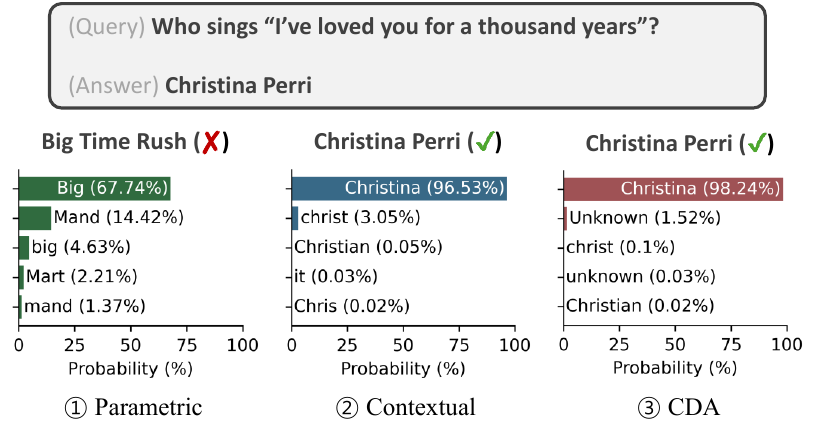} 
        }
        \caption{Output distribution of an answerable query.}
        \label{figure:answerable_distribution}
        
    \end{subfigure}

    \begin{subfigure}{0.95\columnwidth}
        \resizebox{\linewidth}{!}{
            \includegraphics[width=0.94\linewidth]{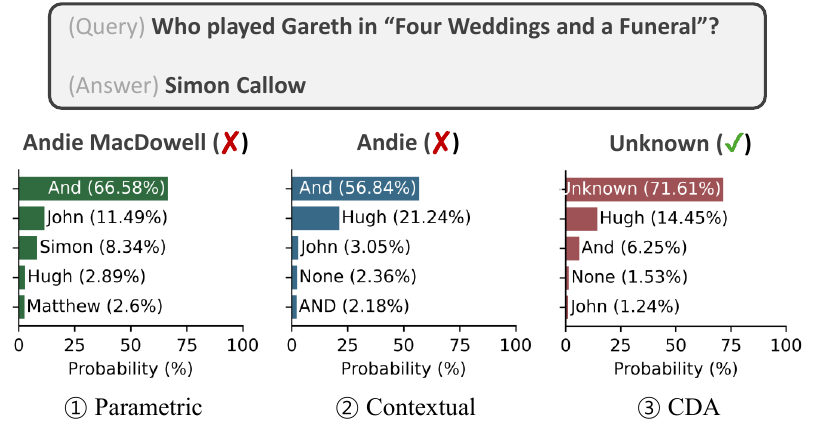} 
        }
        \caption{Output distribution of an unanswerable query.}
        \label{figure:unanswerable_distribution}
        
    \end{subfigure}

    \caption{
        Top-5 probabilities and their corresponding tokens for parametric, contextual, and \method{}  distribution.
        \method{} (a) amplifies the relevant knowledge for answerable queries 
        while (b) shifting the distribution to abstain from unanswerable queries.
    }
    \label{figure:distribution_ablation}
\end{figure}
\subsection{Output Distribution Analysis}

This section analyzes how the output distribution shifts from parametric and contextual distributions to the final distribution of \method{}.
Figure \ref{figure:distribution_ablation} displays the top-5 softmax probabilities along with their corresponding tokens for $d^p_1$, $d^c_1$, and $d^o_1$ from Eq. \ref{eq:ours}.
Figure \ref{figure:answerable_distribution} illustrates an output distribution for an answerable query. 
\method{} successfully attends to the relevant contextual knowledge ``Christina'' and amplifies the probability from 96.53\% to 98.24\%, effectively mitigating the influence of the parametric knowledge.
Furthermore, Figure \ref{figure:unanswerable_distribution} illustrates how \method{} handles unanswerable queries.
While both knowledge erroneously generates the token ``Andie'', \method{} successfully shifts the distribution towards abstention, preventing hallucinations.


\section{Conclusion}
This work addresses the challenge of generating reliable responses leveraging parametric and contextual knowledge when available while abstaining when both are absent.
To evaluate these scenarios, we construct a testbed based on the model's approximated knowledge.
Furthermore, we present \textbf{Contrastive Decoding with Abstention (\method{})}, a novel, training-free decoding method that incorporates abstention in the generation process.
\method{} quantifies the relevance of both knowledge and dynamically attends to the relevant one.
When no relevant knowledge is available, \method{} guides the model to abstain.
Through extensive experiments, \method{} exhibits accurate generation when relevant knowledge is available and abstention otherwise, reducing the risks of hallucination. 


\section*{Limitations}
Our study acknowledges a few limitations that present opportunities for future research.

\paragraph{Computation Cost}
Contrastive decoding inherently involves comparing multiple outputs, inevitably increasing the overall cost.
\method{} also requires additional computations, costing roughly double that of greedy decoding.
However, \method{} enables reliable generation through abstention, a capability that is enhanced through this additional computation.
The reliability of the model may hold greater significance than the inference cost, especially in high-stakes applications. 
We believe that the increased cost is a reasonable trade-off for achieving a more dependable and safe model.
Nonetheless, reducing the overall cost is also an important factor, which will be addressed as a primary objective in our future work.
A detailed analysis of the inference cost is provided in Appendix \ref{appendix:inference_time}.

\paragraph{Limitations in Task Scope}
Our study primarily focuses on single-context scenarios, providing a relatively clear distinction between the presence and absence of knowledge, facilitating precise analysis. 
However, extending the approach to multi-context scenarios would be an important direction for future work. 
Additionally, this work focuses on short-form QA tasks, which are relatively easier to assess the knowledge usage of the model.
However, expanding the task scope reasoning-intensive, long-form generation tasks would be a meaningful advancement.

\paragraph{Advanced Abstention}
The current work mainly focuses on the model's ability to simply express abstention, which lacks user-friendliness.
Future research could explore incorporating reasoning capabilities to explain the rationale behind abstention decisions.

\section*{Acknowledgement}

This work was partly supported by Institute of Information \& communications Technology Planning \& Evaluation (IITP) grant funded by the Korea government (MSIT) [NO.RS-2021-II211343, Artificial Intelligence Graduate School Program (Seoul National University), 
No.RS-2020-II201373, Artificial Intelligence Graduate School Program (Hanyang University), 
NO.RS-2021-II212068, Artificial Intelligence Innovation Hub (Artificial Intelligence Institute, Seoul National University)]
and Mid-Career Bridging Program through Seoul National University.

\bibliography{anthology,custom}

\appendix



\section{Details of Testbed Setting}
\label{appendix:controlled_experimental_setting}
In this section, we provide the details of the testbed setup process.

\subsection{Dataset Details}
\label{appendix:dataset_details}
Natural Questions (NQ) \citep{kwiatkowski-etal-2019-natural}, TriviaQA \citep{joshi-etal-2017-triviaqa}, HotpotQA \citep{yang-etal-2018-hotpotqa} are an open-domain question answering datasets, structured to include a question, a short form answer, and a pre-defined context.
The answer span, including the ground-truth answer, can be found within the context.
Specifically, NQ is composed of information-seeking queries from the Google search engine, and the contexts are Wikipedia pages retrieved by Crowdworkers.
HotpotQA is a multi-hop reasoning dataset consisting of two entity-linked paragraphs from Wikipedia and questions gathered from crowdworkers.
Unlike its original setting, which includes distractor paragraphs, we use the processed version from MRQA \citep{fisch-etal-2019-mrqa}, where the distractors have been removed.
TriviaQA utilizes question-and-answer pairs collected from trivia and quiz-league websites. 
We use the web version of TriviaQA from MRQA.

\subsection{Data Preprocessing}
\label{appendix:data_preprocessing}
The dataset consists of a query $x_i$, a ground-truth answer $y_i$, and a pre-defined context $c_i$.
To keep both inputs and outputs concise, we only utilize the samples with the lengths of the $x_i$ and the $y_i$ limited to 50 and 10 words, respectively.
In cases where $x_i$ appears multiple times within the context $c_i$, we extract multiple corresponding spans ($c_i^1, ..., c_i^k$) for the same query $x_i$.
Each resulting triplets $\{(x_i, c_i^1, y_i), ..., (x_i, c_i^k, y_i)\}$ are included in the dataset $\mathcal{D}_{\text{init}}$.


\paragraph{Relevant Knowledge Estimation}
Although the pre-defined context always contains the answer span, further validation is necessary since it is split into 100-word spans, which may not contain sufficient information to answer the query.
To address this, we measure the sampling consistency by applying a temperature of 1.0 and generating $n=10$ samples for each query.
Each generated sample is compared with the ground-truth answer $y$ to determine its correctness.
The consistency rate is compared with a pre-defined threshold $\eta$ set to 0.7.
The knowledge is considered relevant if the consistency rate exceeds the threshold value.
In other words, the model is deemed to have relevant knowledge for the input $x$ if at least eight samples are correct out of ten generations.

\paragraph{Irrelevant Knowledge Estimation}
For irrelevant context selection,
we utilize SBERT \citep{reimers-gurevych-2019-sentence} embedding to measure the cosine similarity between the training set contexts and the relevant context.
We select the context with the highest cosine similarity as the irrelevant context candidate.
This process is necessary to avoid selecting contexts that are overly unrelated to the query.
Finally, we select contexts with a consistency rate of $r=0$ to ensure they do not provide any unintended information or hints.

\begin{table}[t]
    \centering    
    \resizebox{\linewidth}{!}{


        \begin{tabular}{ccccc}
        \toprule
        Backbone & Seed & NQ & HotpotQA & TriviaQA \\ 
        \midrule
        \multirow{3}{*}{\ml{}} & 1 & 2,338 & 3,614 & 7,524 \\
         & 2 & 1,912 & 3,400 & 9,924 \\
         & 3 & 2,432 & 3,462 & 8,122 \\ 
        \midrule
        \multirow{3}{*}{\mll{}} & 1 & 1,462 & 2,748 & 13,702 \\
         & 2 & 1,758 & 3,538 & 14,034 \\
         & 3 & 2,088 & 3,560 & 14,648 \\ 
        \midrule
        \multirow{3}{*}{\mlll{}} & 1 & 2,150 & 3,508 & 9,806 \\
         & 2 & 2,108 & 3,978 & 10,274 \\
         & 3 & 2,460 & 3,912 & 11,284 \\ 
        \midrule
        \multirow{3}{*}{\mm{}} & 1 & 764 & 1,110 & 9,364 \\
         & 2 & 520 & 1,624 & 11,578 \\
         & 3 & 894 & 1,438 & 11,252 \\ 
         \bottomrule
        \end{tabular}
    
    }
    \caption{
        Number of samples for each dataset constructed for the testbed.
    }
    \label{table:data_count}
\end{table}

\paragraph{Final Dataset Construction} 
To ensure a balanced dataset, we randomly sample
equal number of sample with relevant (i.e., $p_i=1$) and irrelevant (i.e., $p_i=0$) parametric knowledge, matching the size of the smaller set.
The final number of selected samples for each model is presented in Table \ref{table:data_count}.
Note that the number of selected data points varies across models, reflecting the differences in the possessed knowledge.

\subsection{Evaluation Details}
For evaluation, we utilize the final dataset $\text{D}=\{(x_i,c_i^{+},c_i^{-},y_i,p_i)\}^{\text{N}}_{i=1}$.
Specifically, for $x_i \in D$, we evaluate on both input pairs with relevant $(x_i, c^+_i)$ and irrelevant $(x_i, c^-_i)$ contexts.
Only the input $(x_i, c^-_i)$ where $p_i=0$ is deemed unanswerable. 
All other inputs are considered answerable by leveraging either parametric or contextual knowledge.
For example, the input $(x_i, c^+_i)$ where $p_i=0$ is answerable since the model can utilize the relevant context $c^+_i$, even though the model does not pose relevant parametric knowledge (i.e., $p_i=0$).
$(x_i, c^-_i)$ where $p_i=1$ is also answerable since $x_i$ can be answered by utilizing the model's parametric knowledge (i.e., $p_i=1$).
All the experiments are averaged over three different random seeds. 
The full results of \ml{}, \mll{}, \mlll{}, and \mm{} are reported in Table \ref{table:llama3_8b_results}, Table \ref{table:llama2_7b_results}, Table \ref{table:llama2_13b_results}, and Table \ref{table:mistral_7b_results}, respectively.

\section{Experiential Setting Details}
\label{appendix:experiments}
In this section, we describe implementation details for the experiment settings.

\subsection{Implementation Details of \both{}}
\both{} utilize the templates $\mathcal{T}_p(\cdot)$, $\mathcal{T}_c(\cdot)$, and $\mathcal{T}_a(\cdot)$ from Table \ref{template:full}.
For the calibration, we set the query placeholder token $\Bar{x}$ to ``\texttt{[QUESTION]}'' and the context placeholder token $\Bar{c}$ to ``\texttt{[CONTEXT]}''.
The output distribution $\Bar{d}^p_t$ and $\Bar{d}^c_t$ for \nullprompt{} prompts are computed as follows.
\begin{equation}
    \begin{split}
    &\Bar{d}^p_t = \text{logit}_{\theta}(y_t\ |\ \mathcal{T}_p(\Bar{x},y_{<t})), \\    
    &\Bar{d}^c_t = \text{logit}_{\theta}(y_t\ |\ \mathcal{T}_c(\Bar{c},\Bar{x},y_{<t})),
    \end{split}
\end{equation}
For \methodm{}, momentum is applied to each weight as follows:
\begin{equation}
\begin{split}
    &w^c_t \leftarrow \alpha\ w^c_{t-1} + (1-\alpha)\ w^c_{t}, \\
    &w^p_t \leftarrow \alpha\ w^p_{t-1} + (1-\alpha)\ w^p_{t}, \\
    &w^a_t \leftarrow \alpha\ w^a_{t-1} + (1-\alpha)\ w^a_{t},
\end{split}
\end{equation}
where the momentum weight $\alpha$ is set to 0.7.

\subsection{Evaluation Details}
The model is provided with a 2-shot demonstration from the training set and evaluated with the greedy generation.
For answerable queries, the prediction is considered correct if it contains the ground-truth answer \citep{mallen-etal-2023-trust,NEURIPS2023_d842425e}.
Furthermore, the model is expected to appropriately abstain from generating hallucinations for unanswerable queries.
Since there are various ways to abstain, we determine proper abstention by detecting the presence of any pre-defined abstention phrases in the model's output \citep{amayuelas-etal-2024-knowledge,kim-etal-2024-aligning}.
The pre-defined phrases are the following:
[\texttt{unknown answer, answer is unknown, unable to answer, no answer, cannot answer, don't know, do not know}]

\begin{table}[t]
    \centering    

    




    \includegraphics[width=0.95\columnwidth]{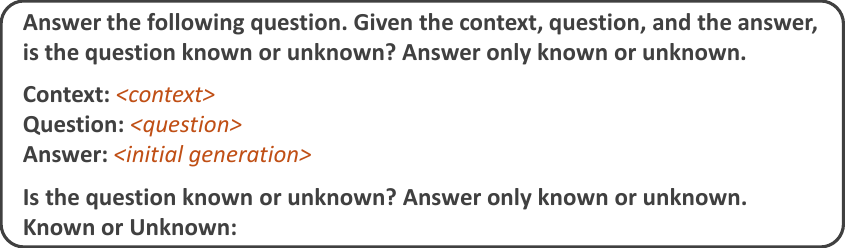}

    \caption{
        Verification template $\mathcal{T}_v(\cdot)$ for \sa{}.
        With the generated answer, original question, and context, 
        the model is prompted to verify whether the question is (un)known.
    }
    \label{template:self_ask}
\end{table}

\subsection{Baselines}
This section provides details of some baselines.


\paragraph{\sa{}} uses $\mathcal{T}_c(\cdot)$ from Table \ref{template:contextual} for the initial generation $\hat{y}$.
The initial generation is "self-asked" to the identical model and is verified using the template $\mathcal{T}_v(c,x,\hat{y})$ from Table \ref{template:self_ask}.
The prediction is abstained if the model generates ``unknown'' as the verification result.

\paragraph{\cad{}} computes the output distribution by amplifying the influence of the contextual knowledge as $d^o_t = d^c_t + w^c_t\ (d^c_t - d^p_t)$.
A fixed weight $ w^c_t$ controls the amount of contextual knowledge applied to the final output distribution.
Following the original work \citep{shi-etal-2024-trusting}, we evaluate the performance with $w^c_t$ set to 0.5 and 1.0 and report the best result.

\paragraph{\ent{}} measures the entropy of the generated tokens when prompted with $\mathcal{T}_c(\cdot)$.
Specifically, for a prediction $\hat{y}=\{\hat{y}^1, ...,  \hat{y}^L,\}$ with $L$-tokens, we leverage the output distribution of each token (i.e., $d^1, ..., d^L$) to measure the token entropy  (i.e., $\mathcal{H}^1, ..., \mathcal{H}^L$) following Eq. \ref{eq:entropy}.
We measure four different variants: first-token ($\mathcal{H}^1$), average ($\frac{1}{L}\sum^L_{i=1}\mathcal{H}^i$), maximum ($\max(\mathcal{H}^1, ..., \mathcal{H}^L)$), and minimum ($\min(\mathcal{H}^1, ..., \mathcal{H}^L)$) entropy.

We compare the entropy measure with a threshold value to perform abstention.
Specifically, if the entropy measure exceeds the threshold value, the prediction is considered uncertain and is abstained.
We utilize the threshold value which demonstrates the best Reliability Score (RS) from the train set.
We report the first-token entropy which yields the best performance as the main result, and the results of other variants are reported only in the Appendix.



\paragraph{\fsb} utilizes $\mathcal{T}_p(\cdot)$, $\mathcal{T}_c(\cdot)$, and $\mathcal{T}_a(\cdot)$ to measure $\text{\entropy}^p_1$, $\text{\entropy}^c_1$, and $\text{\entropy}^a_1$, respectively.
We compare the entropy values at the first decoding step and select the prompting method with the lowest entropy (highest confidence).
For example, if $\text{\entropy}^a_1$ displays the highest confidence, the model continues to generate the same prediction as \abstain{}.


\section{Details on Ablation Experiments}
\label{appendix:ablation}

\begin{figure}[t]
    \begin{center}
        
    \begin{subfigure}{\columnwidth}
        \includegraphics[width=\textwidth]{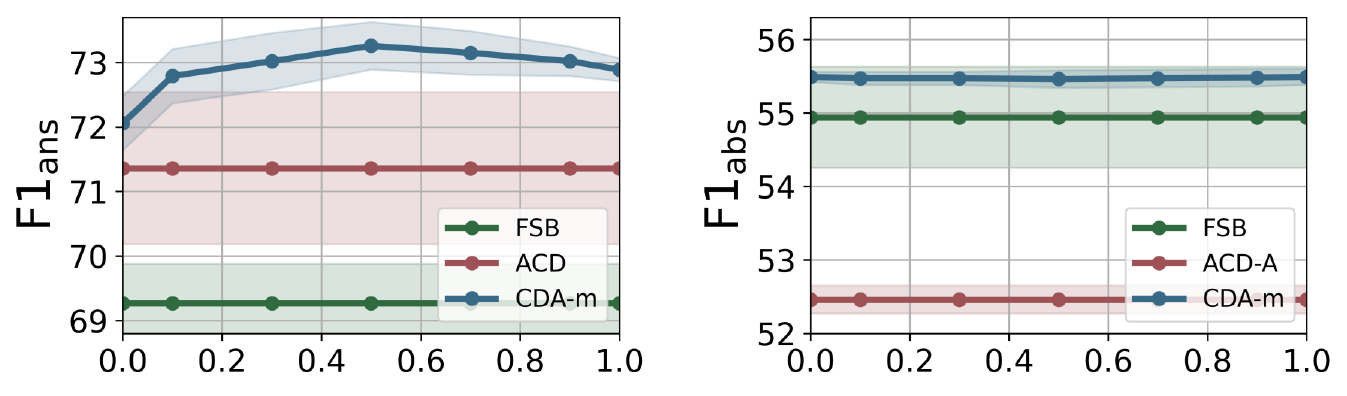} 
        \caption{NQ}
    \end{subfigure}
    
    \begin{subfigure}{\columnwidth}
        \includegraphics[width=\textwidth]{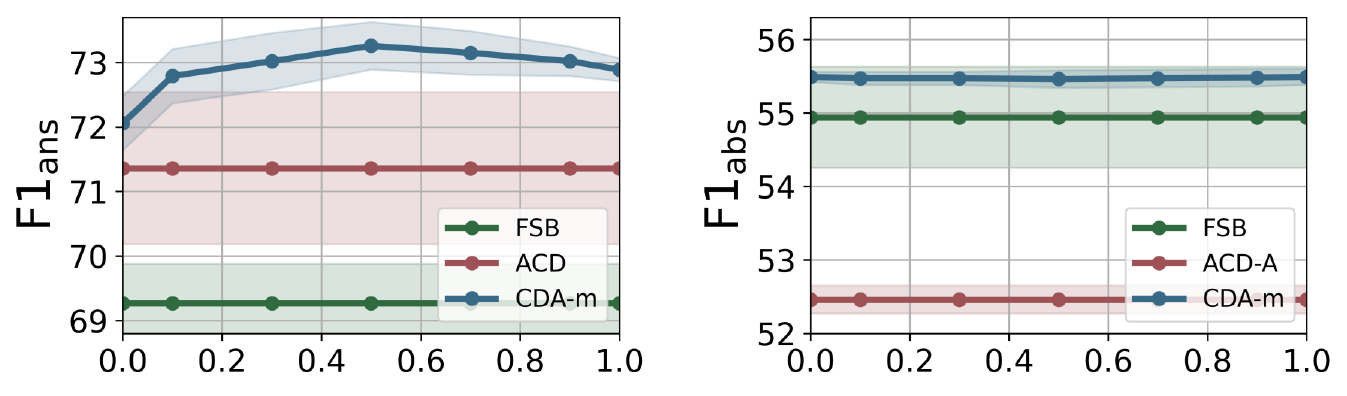} 
        \caption{HotpotQA}
    \end{subfigure}

    \begin{subfigure}{\columnwidth}
        \includegraphics[width=\textwidth]{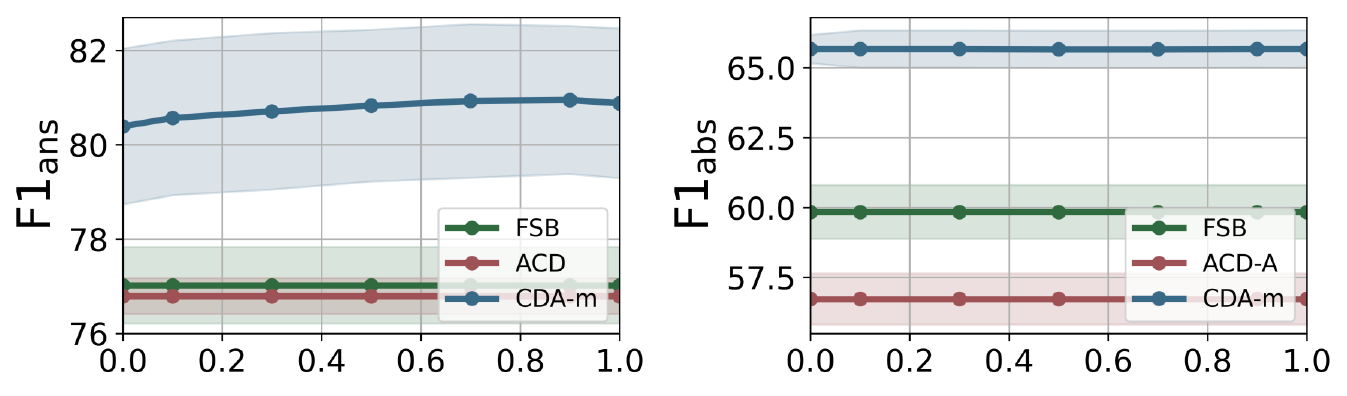} 
        \caption{TriviaQA}
    \end{subfigure}

          \caption{
            \ans{} and \abs{} according to different $\alpha$ values.
            Applying momentum significantly improves \ans{}.
          }
          \label{figure:momentum_full}
    \end{center}
\end{figure}

\subsection{Case Study on Momentum Weight}
\label{appendix:momentum}
\begin{figure}[t!]
    \centering
    
    \begin{subfigure}{\columnwidth}
        
        \resizebox{\linewidth}{!}{
            \includegraphics[width=\linewidth]{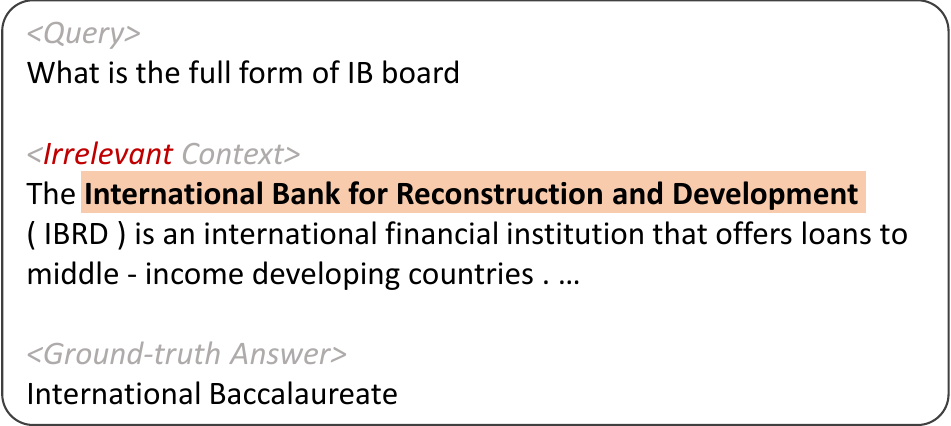} 
        }
        \caption{Input query, irrelevant context, and ground-truth answer.}
        \label{figure:irrelevant_case}
        
    \end{subfigure}

    \begin{subfigure}{\columnwidth}
        
        \resizebox{\linewidth}{!}{
            \includegraphics[width=\linewidth]{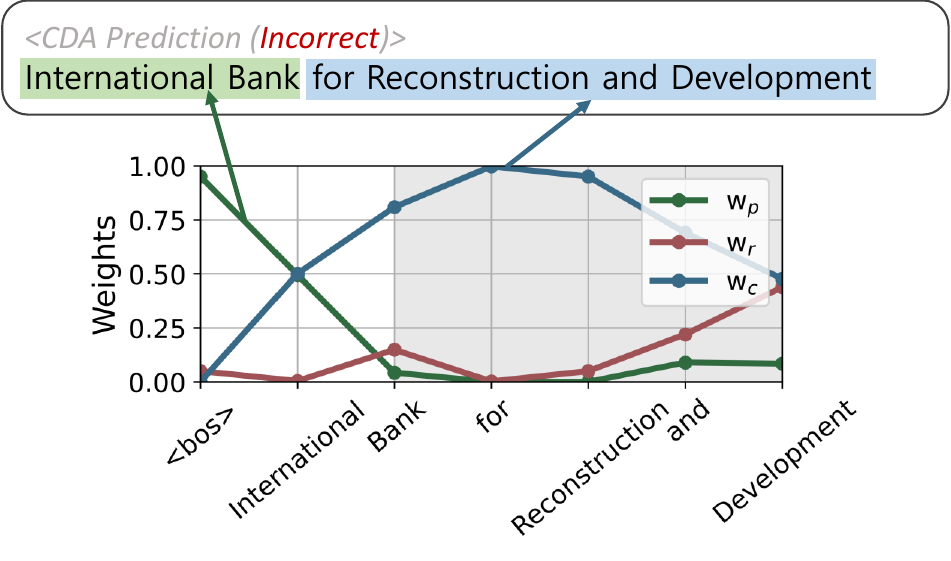} 
        }
        \caption{\method{} output ($\alpha=0.0$)}
        \label{figure:irrelevant_case_cda}
        
    \end{subfigure}

    \begin{subfigure}{\columnwidth}
        
        \resizebox{\linewidth}{!}{
            \includegraphics[width=\linewidth]{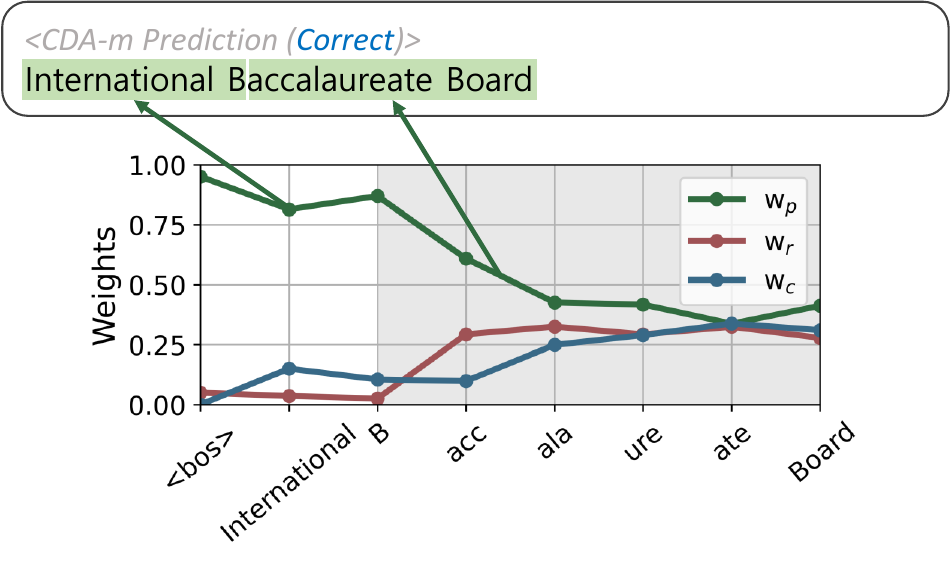} 
        }
        \caption{\methodm{} output ($\alpha=0.7$)}
        \label{figure:irrelevant_case_cdam}
        
    \end{subfigure}

    \caption{
        Example generation of \both{} with an irrelevant context.
        Irrelevant context contains phrases that are similar to the ground-truth answer, making it easier for the model to hallucinate.
        The weights of \method{} shift to the irrelevant context, resulting in incorrect generation.
        On the other hand, the momentum applied to \methodm{} mitigates abrupt shifts in attention, resulting in correct generation.
    }
    \label{figure:case_study_1}
\end{figure}

Table \ref{figure:momentum_full} demonstrates the results of \ans{} and \abs{} with $\alpha$ values from 0.0 to 1.0 in NQ, HotpotQA, and TriviaQA.
The results indicate that incorporating momentum enhances the performance of \ans{}. 
In this section, we conduct a case study regarding this finding.
Figure \ref{figure:irrelevant_case} demonstrates an example where an irrelevant context is provided.
Figure \ref{figure:irrelevant_case_cda} displays the generation result of \method{} and the weights measured for the knowledge at every decoding step. 
\method{} initially assigns more weight to relevant parametric knowledge, generating the correct span up to ``International Bank''.
However, the model's attention shifts toward contextual knowledge, incorporating incorrect information from the context span ``International Bank for Reconstruction and Development.''
We presume that the phrase ``International Bank'' causes this shift in weight, which appears in the irrelevant context.
In contrast, Figure \ref{figure:irrelevant_case_cdam} demonstrates that \methodm{}, leveraging momentum, mitigates abrupt shifts in attention toward irrelevant knowledge, resulting in accurate generation.

\begin{figure}[t!]
    \centering
    
    \begin{subfigure}{\columnwidth}
        
        \resizebox{\linewidth}{!}{
            \includegraphics[width=\linewidth]{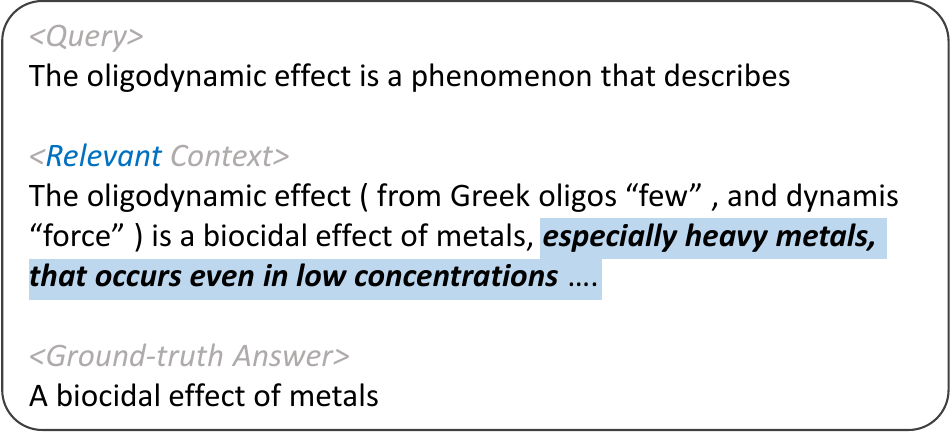} 
        }
        \caption{Input query, relevant context, and ground-truth answer.}
        \label{figure:relevant_case}
        
    \end{subfigure}
    
    \begin{subfigure}{\columnwidth}
        
        \resizebox{\linewidth}{!}{
            \includegraphics[width=\linewidth]{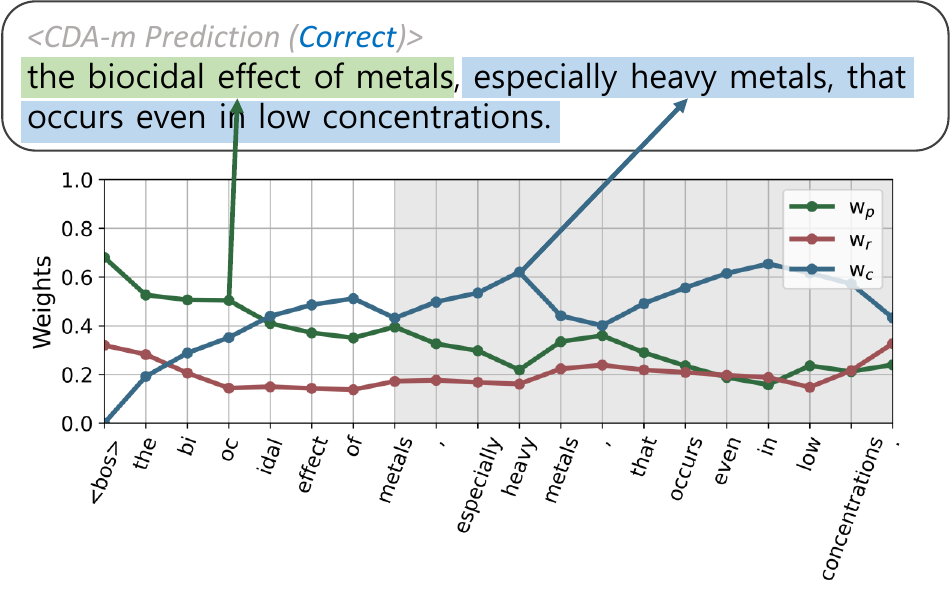} 
        }
        \caption{\methodm{} output ($\alpha=0.7$)}
        \label{figure:relevant_case_cdam}
        
    \end{subfigure}

    \caption{
        Example generation of \methodm{} for a relevant context.
        \methodm{} initially focuses on the parametric knowledge, and the attention shifts to incorporate contextual knowledge, generating richer output.
    }
    \label{figure:case_study_2}
\end{figure}

Notably, \methodm{} does not persistently focus on a single knowledge source.
A case where \methodm{} fully utilizes both knowledge appropriately is shown in Figure \ref{figure:case_study_2}.
Figure \ref{figure:relevant_case} displays a case where a relevant context is provided to the model.
We can observe from Figure \ref{figure:relevant_case_cdam} that \methodm{} initially focuses on the relevant parametric knowledge, and over time, it transitions to incorporate contextual knowledge, producing richer and more nuanced answers.

\subsection{Results of Calibration}
Full ablation results regarding the effect of calibration are reported in Table \ref{table:calibration_ablation_full}.
All the experiments are averaged over three different random seeds.

\begin{table}[t]
    \centering    
    \resizebox{0.85\linewidth}{!}{

        \begin{tabular}{cccccc}
\toprule
Dataset & \abs{} & \ans{} & RS & Acc. & Cov. \\ 
\midrule
NQ & 
\begin{tabular}[c]{@{}c@{}}59.35\\      \footnotesize{(3.43)}\end{tabular} & \begin{tabular}[c]{@{}c@{}}52.06\\      \footnotesize{(1.18)}\end{tabular} & \begin{tabular}[c]{@{}c@{}}47.89\\      \footnotesize{(4.43)}\end{tabular} & \begin{tabular}[c]{@{}c@{}}38.04\\      \footnotesize{(3.87)}\end{tabular} & \begin{tabular}[c]{@{}c@{}}56.79\\      \footnotesize{(2.78)}\end{tabular} \\
HotpotQA & 
\begin{tabular}[c]{@{}c@{}}66.94\\      \footnotesize{(0.84)}\end{tabular} & \begin{tabular}[c]{@{}c@{}}56.39\\      \footnotesize{(0.21)}\end{tabular} & \begin{tabular}[c]{@{}c@{}}56.42\\      \footnotesize{(1.33)}\end{tabular} & \begin{tabular}[c]{@{}c@{}}45.43\\      \footnotesize{(1.29)}\end{tabular} & \begin{tabular}[c]{@{}c@{}}63.55\\      \footnotesize{(0.69)}\end{tabular} \\
TriviaQA & 
\begin{tabular}[c]{@{}c@{}}67.56\\      \footnotesize{(1.27)}\end{tabular} & \begin{tabular}[c]{@{}c@{}}57.24\\      \footnotesize{(0.63)}\end{tabular} & \begin{tabular}[c]{@{}c@{}}56.46\\      \footnotesize{(1.19)}\end{tabular} & \begin{tabular}[c]{@{}c@{}}45.32\\      \footnotesize{(1.14)}\end{tabular} & \begin{tabular}[c]{@{}c@{}}64.17\\      \footnotesize{(0.99)}\end{tabular} \\ 
\bottomrule
\end{tabular}

    }
    \caption{
        Average and standard deviation (in parentheses) of CDA without calibration. 
    }
    \label{table:calibration_ablation_full}
\end{table}


\subsection{Implementation Details of Training-based Methods}
In this section, we describe the implementation details of training-based methods.
Besides \textbf{instruction-tuning}, we also utilize \textbf{external verifier} \citep{cobbe2021training,cohen2023crawling} for comparison.
We report the average performance over three different random seeds.
The full results are displayed in Table \ref{table:trained_methods_full}.

\paragraph{External Verifier}
We utilize RoBERTa-base as an external verifier. The overall training process is as follows.
First, we prompt the inference model (e.g., \ml{}) with the prompt $\mathcal{T}_c(c,x)$ and generate a prediction $\hat{y}$.
Then, we measure the correctness of $\hat{y}$ by comparing it with the ground-truth answer $y$.
The label for the verifier $\Bar{y}$ is assigned based on the correctness of $\hat{y}$.
Specifically, we assign $\Bar{y}=1$ when $\hat{y}$ is correct and $\Bar{y}=0$ when $\hat{y}$ is incorrect.
We feed $\mathcal{T}_c(c,x,\hat{y})$ into the verifier and pass the \texttt{[CLS]} token through a single MLP layer for the final prediction.
The verifier is trained to predict $\Bar{y}$ given $\mathcal{T}_c(c,x,\hat{y})$.
During inference, we first generate $\hat{y}$ from the inference model.
Then, we pass $\mathcal{T}_c(c,x,\hat{y})$ to the verifier to predict the correctness of the prediction $\hat{y}$.
Samples classified as incorrect by the verifier are abstained.

\definecolor{bg}{HTML}{b1d4e0}

\begin{table}[t]
    \centering
        
    \begin{subfigure}{0.99\columnwidth}
        
        \resizebox{\linewidth}{!}{

        \begin{tabular}{ccccc}
        \toprule
        Source & Target & \abs{} & \ans{} & RS \\ 
        \midrule
        \multirow{3}{*}{NQ} & 
        \cellcolor{bg}NQ & 
        \cellcolor{bg}64.86 \footnotesize{(1.49)} &
        \cellcolor{bg}46.56 \footnotesize{(3.17)} & 
        \cellcolor{bg}58.03 \footnotesize{(1.06)} \\
         & HotpotQA & 66.62 \footnotesize{(1.90)} & 52.36 \footnotesize{(2.61)} & 58.50 \footnotesize{(1.16)} \\
         & TriviaQA & 65.07 \footnotesize{(2.09)} & 53.52 \footnotesize{(3.07)} & 57.39 \footnotesize{(2.17)} \\ 
        \midrule
        \multirow{3}{*}{HotpotQA} & NQ & 61.46 \footnotesize{(3.13)} & 38.42 \footnotesize{(8.06)} & 54.60 \footnotesize{(2.15)} \\
         & \cellcolor{bg}HotpotQA & 
         \cellcolor{bg}68.41 \footnotesize{(0.22)} & 
         \cellcolor{bg}56.92 \footnotesize{(1.32)} & 
         \cellcolor{bg}60.80 \footnotesize{(1.12)} \\
         & TriviaQA & 60.09 \footnotesize{(2.37)} & 38.53 \footnotesize{(11.96)} & 53.28 \footnotesize{(1.04)} \\ 
        \midrule
        \multirow{3}{*}{TriviaQA} & NQ & 63.30 \footnotesize{(0.74)} & 45.10 \footnotesize{(5.07)} & 55.55 \footnotesize{(0.27)} \\
         & HotpotQA & 66.97 \footnotesize{(0.19)} & 54.88 \footnotesize{(0.98)} & 59.24 \footnotesize{(0.26)} \\
         & \cellcolor{bg}TriviaQA & 
         \cellcolor{bg}65.31 \footnotesize{(1.14)} & 
         \cellcolor{bg}52.63 \footnotesize{(2.56)} & 
         \cellcolor{bg}58.13 \footnotesize{(0.43)} \\ 
         \bottomrule
        \end{tabular}
        }
        \caption{Results of external verifier.}
        
    \end{subfigure}

    \begin{subfigure}{0.99\columnwidth}
        \resizebox{\linewidth}{!}{

        \begin{tabular}{ccccc}
        \toprule
        Source & Target & \ans{} & \abs{} & RS \\ 
        \midrule
        \multirow{3}{*}{NQ} & 
        \cellcolor{bg}NQ & 
        \cellcolor{bg} 66.37 \footnotesize{(0.55)} &
        \cellcolor{bg} 43.87 \footnotesize{(2.02)} &
        \cellcolor{bg} 61.12 \footnotesize{(0.46)} \\
         & HotpotQA &  
         64.43 \footnotesize{(0.42)} &  
         48.79 \footnotesize{(3.88)} &
         59.65 \footnotesize{(0.42)} \\
         & TriviaQA & 
          67.31 \footnotesize{(1.82)} &
          49.00 \footnotesize{(4.74)} &
          62.14 \footnotesize{(1.43)} \\ 
         \midrule
         
        \multirow{3}{*}{HotpotQA} & 
        NQ &  
        64.86 \footnotesize{(2.38)} &
        42.73 \footnotesize{(1.49)} &  
        59.80 \footnotesize{(1.92)} \\
         & 
         \cellcolor{bg}HotpotQA & 
         \cellcolor{bg} 74.06 \footnotesize{(0.76)} &
         \cellcolor{bg} 57.57 \footnotesize{(0.73)} &
         \cellcolor{bg} 68.76 \footnotesize{(0.81)} \\
         & TriviaQA &  
         65.26 \footnotesize{(4.92)} &
         50.69 \footnotesize{(2.06)} &
         61.17 \footnotesize{(4.10)} \\ 
         
        \midrule
        \multirow{3}{*}{TriviaQA} & 
        NQ &  
        65.15 \footnotesize{(1.66)} &
        46.37 \footnotesize{(4.00)} &
        59.70 \footnotesize{(1.10)} \\
         & HotpotQA &  
         66.68 \footnotesize{(0.13)} &
         49.82 \footnotesize{(3.52)} &
         60.58 \footnotesize{(0.25)} \\
         & 
         \cellcolor{bg}TriviaQA &
         \cellcolor{bg} 67.60 \footnotesize{(1.73)} &
         \cellcolor{bg} 53.68 \footnotesize{(2.92)} &
         \cellcolor{bg} 61.84 \footnotesize{(0.70)} \\ 
         \bottomrule
        \end{tabular}
        }
        \caption{Results of instruction-tuning.}
        
    \end{subfigure}
    
    \caption{
        Average and standard deviation (in parentheses) of training-based methods over three different random seeds.
        \colorbox{bg}{In-domain} results are highlighted in blue.
    }
    \label{table:trained_methods_full}
\end{table}

\paragraph{Instruction-tuning}
For instruction-tuning, we first re-label the training data based on the model's knowledge.
Specifically, we follow the same knowledge estimation process from Section \ref{sec:controlled_exp} and estimate the parametric and contextual knowledge of the given training sample.
If the model possesses at least one relevant knowledge for the given $x$, it is labeled with the ground-truth answer $y$.
However, if the model does not have any relevant knowledge, $x$ is labeled with a pre-defined abstention response $y_{\text{abs}}$ (e.g., ``unknown'').
The model is then trained to generate the label given $\mathcal{T}_c(c,x)$.
For evaluation, we utilize the instruction-tuned model for the prediction with $\mathcal{T}_c(c,x)$ as the input.

\paragraph{Training Details}
For instruction-tuning, we employ QLoRA \citep{NEURIPS2023_1feb8787} from Huggingface PEFT library \citep{peft} with $r=4$ and $alpha=16$ for efficient training.
We select the model with the best Reliability Score (RS) performance in the validation set from learning rates 
\texttt{[1e-3, 5e-4, 1e-4, 5e-5, 1e-5, 5e-6, 1e-6]} and training epochs \texttt{[2, 3, 4, 5]}.
All results are averaged over three different random seeds. 
Figure \ref{table:trained_methods_full} displays the full results of the external verifier and instruction-tuning in both in-domain (IND) and out-of-domain (OOD) scenarios.
Overall, instruction-tuning displays better performance than the external verifier in IND scenarios. 
However, both methods exhibit a \textbf{notable degradation in OOD inputs}, limiting their generalizability.


\begin{table}[t]
    \centering    
    \resizebox{\linewidth}{!}{

    \begin{tabular}{cccc}
    \toprule
    Dataset & \context{} & \method{} & \methodm{} \\ 
    \midrule
    NQ & 258.69 & 622.81 \footnotesize{($\times$ 2.41)} & 645.78 \footnotesize{($\times$ 2.50)} \\ 
    HotpotQA & 194.87 & 486.82 \footnotesize{($\times$ 2.50)} &  535.67 \footnotesize{($\times$ 2.75)} \\ 
    TriviaQA & 188.92 & 482.65 \footnotesize{($\times$ 2.55)} &  461.87 \footnotesize{($\times$ 2.44)} \\ 
    \bottomrule
    \end{tabular}

    }
    \caption{
        Total computation time (in seconds) for \ml{} generating 100 samples.
        We can observe that the total computation time of \both{} is roughly two times that of \context{}.
    }
    \label{table:inference_time}
\end{table}


\subsection{Evaluation on RAG Setting}
Table \ref{table:rag_full} presents the Reliability Score (RS) results across all the datasets and backbones.
Results of \ent{} variants (average, maximum, minimum entropy) are also reported.
\both{} consistently outperform the baselines, underscoring their effectiveness in practical scenarios.

\begin{table}[t!]
    \centering    
    \resizebox{0.93\linewidth}{!}{

    \footnotesize
        \begin{tabular}{cccc}
\toprule
Method & NQ & HotpotQA & TriviaQA \\ 
\midrule
\multicolumn{4}{l}{\ml{}} \\ 
\midrule
\context{} & 33.35 & 31.00 & 65.30 \\
\cad{} & 28.86 & 28.24 & 59.32 \\
\acd{} & 38.40 & 34.05 & 73.91 \\
\abstain{} & 50.74 & 50.71 & 76.66 \\
\sa{} & 45.67 & 39.61 & 64.91 \\
\ent{} (\textit{first-token}) & 50.84 & 50.24 & 76.20 \\
\ent{} (\textit{average}) & 48.46 & 47.67 & 72.12 \\
\ent{} (\textit{max}) & 49.40 & 48.34 & 74.24 \\
\ent{} (\textit{min}) & 44.76 & 43.58 & 68.61 \\
\fsb{} & 53.66 & 51.07 & 79.14 \\
\acda{} & 51.80 & 50.35 & 75.58 \\ 
\midrule
\method{} & \textbf{54.33} & \underline{51.77} & \underline{80.62} \\
\methodm{} & \underline{54.32} & \textbf{51.80} & \textbf{80.67} \\ 
\midrule
\multicolumn{4}{l}{\mll{}} \\ 
\midrule
\context{} & 30.11 & 28.28 & 61.01 \\
\cad{} & 27.20 & 25.86 & 54.44 \\
\acd{} & 33.02 & 30.04 & 67.35 \\
\abstain{} & 22.06 & 20.02 & 70.70 \\
\sa{} & 31.72 & 33.99 & 63.12 \\
\ent{} (\textit{first-token}) & 41.94 & 44.71 & 71.87 \\
\ent{} (\textit{average}) & 42.88 & 45.01 & 70.12 \\
\ent{} (\textit{max}) & 41.36 & \textbf{45.69} & 71.88 \\
\ent{} (\textit{min}) & 41.30 & 41.65 & 65.67 \\
\fsb{} & 37.36 & 39.58 & 73.13 \\
\acda{} & 33.56 & 35.05 & 70.55 \\ 
\midrule
\method{} & \textbf{47.09} & \underline{45.07} & \underline{73.18} \\
\methodm{} & \underline{46.40} & 45.02 & \textbf{73.90} \\ 
\midrule
\multicolumn{4}{l}{\mlll{}} \\ 
\midrule
\context{} & 31.15 & 30.57 & 66.81 \\
\cad{} & 28.40 & 27.59 & 61.80 \\
\acd{} & 35.65 & 32.73 & 72.09 \\
\abstain{} & 41.54 & 24.54 & 47.11 \\
\sa{} & 34.15 & 33.68 & 67.73 \\
\ent{} (\textit{first-token}) & 48.28 & 44.31 & 74.56 \\
\ent{} (\textit{average}) & 47.63 & 43.31 & 73.88 \\
\ent{} (\textit{max}) & 48.61 & 43.93 & 75.11 \\
\ent{} (\textit{min}) & 43.73 & 39.88 & 70.18 \\
\fsb{} & 49.88 & \textbf{45.28} & 72.79 \\
\acda{} & 48.30 & 41.62 & 65.78 \\ 
\midrule
\method{} & \underline{51.42} & 44.06 & \underline{77.89} \\
\methodm{} & \textbf{51.95} & \underline{44.73} & \textbf{78.74} \\ 
\midrule
\multicolumn{4}{l}{\mm{}} \\ 
\midrule
\context{} & 30.68 & 29.00 & 62.40 \\
\cad{} & 28.60 & 26.08 & 56.09 \\
\acd{} & 32.42 & 28.91 & 66.24 \\
\abstain{} & 49.11 & 41.99 & \underline{73.00} \\
\sa{} & 39.34 & 38.78 & 67.94 \\
\ent{} (\textit{first-token}) & 48.65 & 46.82 & 71.69 \\
\ent{} (\textit{average}) & 46.84 & 46.69 & 71.54 \\
\ent{} (\textit{max}) & 48.38 & 47.06 & 71.57 \\
\ent{} (\textit{min}) & 42.58 & 42.76 & 66.19 \\
\fsb{} & 49.05 & \textbf{47.32} & 70.26 \\
\acda{} & 48.42 & 47.14 & 70.39 \\ 
\midrule
\method{} & \underline{49.36} & 46.94 & 72.86 \\
\methodm{} & \textbf{49.51} & \underline{47.13} & \textbf{73.22} \\ 
\bottomrule
\end{tabular}

    }
    \caption{
        Results of Reliability Score (RS) across all the scenarios in the RAG setting.
        \both{} consistently outperform all the baselines.
    }
    \label{table:rag_full}
\end{table}


\section{Computation Cost Analysis}
\label{appendix:inference_time}
Contrastive decoding enables the model to utilize various knowledge and abilities by leveraging multiple output distributions. 
However, the process incurs additional costs, which is also an inherent limitation of \method{}.
In this section, we analyze the computation cost of \method{} in detail.

Let the lengths of the template, context, and query be $L_t$, $L_c$, and $L_q$, respectively.
The computational cost for a single inference (e.g., \context{}) is proportional to the square of the total input length $O((L_t + L_c + L_q)^2)$.
\method{} performs five inferences of contextual prompting $O((L_t + L_c + L_q)^2)$,
parametric prompting $O((L_t + L_q)^2)$,
abstention prompting $O((L_t + L_c + L_q)^2)$,
contextual \nullprompt{} prompting $O((L_t + 2)^2)$, and
parametric \nullprompt{} prompting $O((L_t + 1)^2)$.
While this may appear computationally heavy, in practice, the template and the query are relatively short, while the context is typically long, making the overall computation roughly $2 * O(L_c^2)$.
Consequently, the additional cost required in \method{} is \textbf{approximately twice that of a single inference}.

To validate these estimates, we conduct an experiment on \ml{} using 100 randomly selected samples from all the datasets.
We compare the total generation time of \context{} with \method{} in seconds on a single RTX 3090 GPU. 
Table \ref{table:inference_time} demonstrates the overall inference time.
Empirical results indicate that the inference time of \method{} does not exceed three times that of a single inference.

\definecolor{my_gray}{HTML}{E8E8E8}
\newcolumntype{g}{>{\columncolor{my_gray}}c}

\begin{table*}[t!]
    \centering    
    \resizebox{0.99\textwidth}{!}{


    }
    \caption{
        Average and standard deviation (in parentheses) of the results over three different random seeds in \ml{}. 
        The \textbf{best result} is highlighted in bold, and the \underline{second-best result} is underlined.
    }
    \label{table:llama3_8b_results}
\end{table*}
\definecolor{my_gray}{HTML}{E8E8E8}
\newcolumntype{g}{>{\columncolor{my_gray}}c}

\begin{table*}[t!]
    \centering    
    \resizebox{0.99\textwidth}{!}{
%

    }
    \caption{
        Average and standard deviation \footnotesize{(in parentheses)} of the results over three different random seeds in \mll{}. 
        The \textbf{best result} is highlighted in bold, and the \underline{second-best result} is underlined.
    }
    \label{table:llama2_7b_results}
\end{table*}
\definecolor{my_gray}{HTML}{E8E8E8}
\newcolumntype{g}{>{\columncolor{my_gray}}c}

\begin{table*}[t!]
    \centering    
    \resizebox{0.99\textwidth}{!}{
%

    }
    \caption{
        Average and standard deviation \footnotesize{(in parentheses)} of the results over three different random seeds in \mlll{}. 
        The \textbf{best result} is highlighted in bold, and the \underline{second-best result} is underlined.
    }
    \label{table:llama2_13b_results}
\end{table*}
\definecolor{my_gray}{HTML}{E8E8E8}
\newcolumntype{g}{>{\columncolor{my_gray}}c}

\begin{table*}[t!]
    \centering    
    \resizebox{0.99\textwidth}{!}{
%
 \\ 
\bottomrule
\end{tabular}

    }
    \caption{
        Average and standard deviation \footnotesize{(in parentheses)} of the results over three different random seeds in \mm{}. 
        The \textbf{best result} is highlighted in bold, and the \underline{second-best result} is underlined.
    }
    \label{table:mistral_7b_results}
\end{table*}

\end{document}